\documentclass[lettersize,journal]{IEEEtran}
\usepackage{amsmath,amsfonts,amssymb,amsthm}
\usepackage{algorithmic}
\usepackage{algorithm}
\usepackage{array}
\usepackage{textcomp}
\usepackage{stfloats}
\usepackage{url}
\usepackage{verbatim}
\usepackage{graphicx}
\usepackage{cite}
\usepackage{subfig}
\usepackage{color}
\usepackage{colortbl}
\usepackage{soul}
\usepackage{multirow}
\usepackage{float}

\usepackage{pifont}% http://ctan.org/pkg/pifont

\newcommand{\etal}{\textit{et al}.}

\usepackage{longtable}
\usepackage{booktabs}
\usepackage[table]{xcolor}%
\definecolor{LightRed}{rgb}{1,0.92,0.92}
\definecolor{LightOrange}{rgb}{1,0.95,0.88}
\definecolor{LightYellow}{rgb}{1.0,1.0,0.84}
\definecolor{LightGreen}{rgb}{0.9,1.0,0.88}
\definecolor{LightCyan}{rgb}{0.9,1,1}
\definecolor{LightBlue}{rgb}{0.9,0.94,1}
\definecolor{LightIndigo}{rgb}{0.92,0.9,1}
\definecolor{LightMagenta}{rgb}{0.96,0.86,1}
\definecolor{DirtyWhite}{rgb}{0.96,0.96,0.96}
\definecolor{almond}{rgb}{0.94, 0.87, 0.8}
\definecolor{babypink}{rgb}{0.96, 0.76, 0.76}
\definecolor{blond}{rgb}{0.98, 0.94, 0.75}

\newtheorem{theorem}{Theorem}%  meant for continuous numbers
\newtheorem{definition}{Definition}%

\usepackage{tabularx}
\usepackage{makecell}
\usepackage{adjustbox}
\usepackage{booktabs}%

\newcommand{\rotbox}[1]{\rotatebox{90}{#1}}
\definecolor{tabhighlight}{HTML}{e5e5e5}
\begin{document}

\title{A Survey of Low-shot Vision-Language Model Adaptation via Representer Theorem}

\author{Kun Ding, Ying Wang, Gaofeng~Meng, Shiming Xiang
	% <-this % stops a space
	\thanks{K. Ding, Y. Wang, G. Meng and S. Xiang are with the State Key Laboratory of Multimodal Artificial Intelligence Systems (MAIS), 
		Institute of Automation, Chinese Academy of Sciences, Beijing 100190, China (e-mail: kun.ding@ia.ac.cn, ywang@nlpr.ia.ac.cn, gfmeng@nlpr.ia.ac.cn, smxiang@nlpr.ia.ac.cn).}% <-this % stops a space
}

% The paper headers
\markboth{Journal of \LaTeX\ Class Files,~Vol.~14, No.~8, August~2021}%
{Shell \MakeLowercase{\textit{et al.}}: A Sample Article Using IEEEtran.cls for IEEE Journals}

%\IEEEpubid{0000--0000/00\$00.00~\copyright~2021 IEEE}
% Remember, if you use this you must call \IEEEpubidadjcol in the second
% column for its text to clear the IEEEpubid mark.

\maketitle

\begin{abstract}
The advent of pre-trained vision-language foundation models has revolutionized the field of zero/few-shot (i.e., low-shot) image recognition. The key challenge to address under the condition of limited training data is how to fine-tune pre-trained vision-language models in a parameter-efficient manner. Previously, numerous approaches tackling this challenge have been proposed. Meantime, a few survey papers are also published to summarize these works. However, there still lacks a unified computational framework to integrate existing methods together, identify their nature and support in-depth comparison. As such, this survey paper first proposes a unified computational framework from the perspective of Representer Theorem and then derives many of the existing methods by specializing this framework. Thereafter, a comparative analysis is conducted to uncover the differences and relationships between existing methods. Based on the analyses, some possible variants to improve the existing works are presented. As a demonstration, we extend existing methods by modeling inter-class correlation between representers in reproducing kernel Hilbert space (RKHS), which is implemented by exploiting the closed-form solution of kernel ridge regression. Extensive experiments on 11 datasets are conducted to validate the effectiveness of this method. Toward the end of this paper, we discuss the limitations and provide further research directions.
\end{abstract}


\section{Introduction}
\label{sec:introduction}
\IEEEPARstart{L}{arge} language models (LLMs)~\cite{GPT,LLaMA} have revolutionized the field of natural language processing (NLP), exhibiting remarkable proficiency across a broad spectrum of NLP tasks. However, LLMs only process single input modality, i.e., text, limiting their application scope. Recent advancements are centered on the development of large vision-language models (LVLMs)~\cite{CLIP,BLIP,Flamingo,LLaVa}, which integrate multi-modal inputs, including image and text, thereby enabling a more extensive range of applications, such as cross-modal retrieval~\cite{CLIP}, image captioning~\cite{Flamingo}, visual question answering~\cite{LLaVa}, text-guided generation~\cite{ControlNet}. Within the realm of LVLMs, contrastive vision-language models (CVLMs)~\cite{CLIP,ALIGN,SLIP,CuiZGYWY022,ALBEF,alphaCLIP} like CLIP~\cite{CLIP} and ALIGN~\cite{ALIGN} pre-trained on web-scale noisy image-text pairs by optimizing cross-modal constrastive learning loss have gained widespread attention due to their strong ability as foundation models. Demonstrated by recent works~\cite{CLIP,CoOp,DenseCLIP}, CVLMs are capable of extracting image features rich in semantic information and exhibiting strong generalization abilities. For this aim, CVLMs have been widely used in image segmentation~\cite{DenseCLIP,ZegCLIP,CRIS}, open-vocabulary object detection~\cite{CORA,EdaDet}, zero/few-shot (i.e., low-shot) image recognition~\cite{CoOp,TipAdapter,CoCoOp,APE}, and so on.

Among these tasks, low-shot image recognition stands as a pivotal task in computer vision (CV), aiming to recognize images with only a few or no labeled training samples. This task has obtained extensive researches and lots of methods have been developed from diverse perspectives~\cite{Song0CMS23}, encompassing data augmentation, transfer learning, meta learning, and so on. Despite these efforts, this task still faces many challenges. Fortunately, recent advancements in foundation vision-language models have shed light on new directions for solving the low-shot recognition problem.

Recent methods try to solve the low-shot image recognition problem by adapting pre-trained foundation vision-language models, especially the CVLMs. The key challenge for this aim is how to implement parameter-efficient fine-tuning (PEFT)~\cite{PEFTforLM_survey,PEFTforVision,XingLWSCGW24}. Traditional tuning methods, such as full-parameter fine-tuning, can easily destroy the pre-trained features, leading to sub-optimal generalizability. By contrast, PEFT only tunes a small portion of the original parameters or a few newly attached parameters, achieving better data efficiency and generalization ability.

Recent years have witnessed various PEFT methods for adapting CVLMs to low-shot recognition tasks. These methods can broadly be classified into two categories: prompt-based~\cite{CoOp,CoCoOp,KgCoOp,MaPLe,WeakDistributionDet,LASP} and adapter-based methods~\cite{CLIP_Adapter,TipAdapter,APE,CPR}. The prompt-based methods involve adding some context tokens to the input sequence of text encoder (i.e., the language model), and the tokens are usually made learnable. The learnable tokens enhance the plasticity of pre-trained models compared to non-learnable ones, thereby more suitable for few-shot learning. The non-learnable ones avoid using extra training samples, which are more suitable for zero-shot learning. As for adapter-based methods, deeper image or text features instead of the shallower ones are tuned, which are usually achieved by inserting tiny networks or layers in the last several layers and only optimizing their parameters. More recent adapter-based methods~\cite{TipAdapter,APE} usually introduce a cache model to enhance logits. The cache model exploits the information from training images, acting as a non-parametric classifier.

The existing methods for adapting pre-trained CVLMs to low-shot image recognition tasks vary widely in terms of the position of the tunable module inserted, the count of the parameters being tuned and the computational process, making grasping the underlying idea of them even harder. Consequently, it is urgent to summarize and organize these methods, analyze their differences and connections, and provide insights for further development directions. Considering the relation between PEFT and the Representer Theorem for regularized empirical risk minimization in reproducing kernel Hilbert space (RKHS), we propose a unified computational framework for PEFT of CVLMs, which is consisted of \textbf{five} components: \textbf{input image encoding, anchors computation, kernels computation, logits computation and loss function computation}. The similarities and differences of existing methods can be more clearly compared in this framework. Based on the comparison, we present some possible variants of existing methods within our framework to further enhance their performance. Among the proposed variants, we demonstrate how to exploit the closed-form solution of kernel ridge regression (KRR) to model inter-class correlation between representers (i.e., functions in reproducing kernel Hilbert space) in the logits computation step. The effectiveness of the proposed method is validated through extensive experiments on 11 public datasets. Toward the conclusion of this paper, we delve into the limitations of our framework and propose prospective research directions.

In summary, the contributions of this work are listed as follows:
\begin{itemize}
	\item A unified computational framework based on the Representer Theorem is proposed to integrate many of the existing methods together. Within the framework, the differences and relations between these methods are clarified from five dimensions, which can help the researchers to strengthen the understanding of previous PEFT methods.
	\item Several possible variants improving existing methods within our framework are proposed, including modeling inter-class relation of representers, exploiting multi-kernel learning, designing more fine-grained kernel functions, and so on. The exploration of these variants is expected to enhance the performance of existing PEFT methods for low-shot image recognition.
	\item A demonstrative implementation of modeling inter-class relation between representers is proposed. The effectiveness of the method is validated by experiments on 11 public datasets. The successful demonstration of the effectiveness implies that it is possible to obtain some inspirational ideas in our framework that can improve existing works, showcasing its value.
\end{itemize}

In the literature, there already exist some works that survey the PEFT methods. For example, Han \etal~\cite{PEFTforLM_survey} proposed a comprehensive review for PEFT methods for large models, including five aspects: background, taxonomy, design, application and challenge. However, they mainly summarized the methods for tuning language models. Xin \etal~\cite{PEFTforVision} provided a review of PEFT for pre-trained vision models. The models covered include supervised pre-trained vision models (e.g., SAM~\cite{SAM}) and self-supervised pre-trained vision models (e.g., CLIP~\cite{CLIP}, DINO~\cite{DINO}). Xing \etal~\cite{XingLWSCGW24} surveyed efficient fine-tuning methods for vision-language models, but not dedicated to CVLMs and low-shot recognition task. In contrast to these works, this survey does not aim to provide a holistic overview of PEFT methods for general tasks and vision-language models. Instead, we seek to propose a unified framework or architecture specifically tailored for tuning CVLMs for low-shot image recognition. By narrowing our focus, we aim to deliver a focused and in-depth survey of PEFT methods for CVLMs in the context of low-shot image recognition.

\begin{figure*}[!thb]
	\centering
	\includegraphics[width=0.82\linewidth]{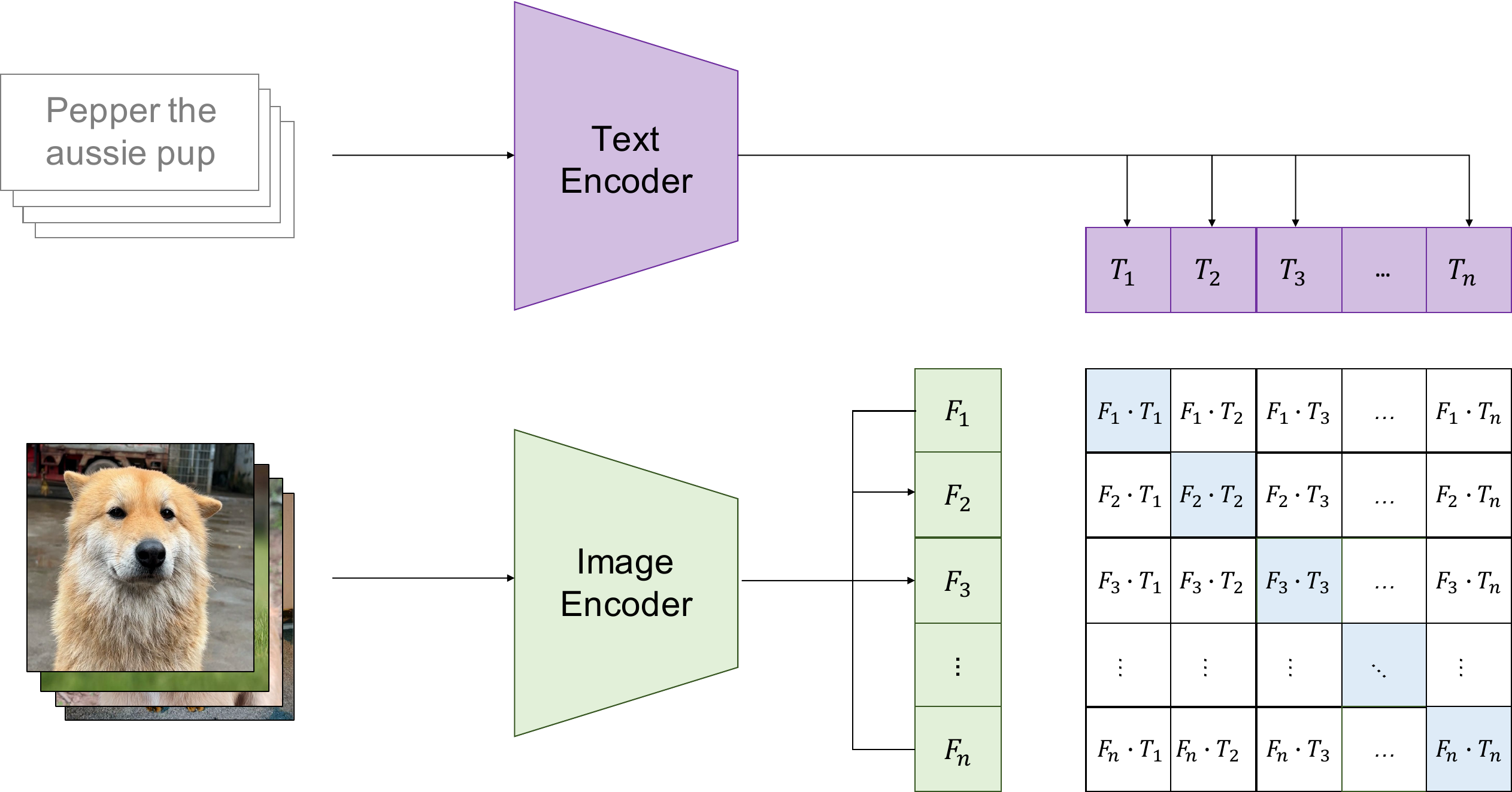}
	\caption{Overview of CLIP (modified from \cite{CLIP}). In this figure, $F_1, \cdots, F_n$ are the image features extracted by the image encoder, $T_1, \cdots, T_n$ are the text features extracted by the text encoder.}
	\label{fig:CLIP}
\end{figure*}

The rest of this paper will be organized in the following order: 1) In Section~\ref{sec:pre}, we introduce some background knowledge about CVLMs and give formal definition of the involved problem. 2) In Section~\ref{sec:framework}, we will present the proposed framework. We first give a concise introduction to the Representer Theorem and then apply it to proposing the framework. 3) In Section~\ref{sec:instantiations}, we reproduce existing PEFT methods by instantiating the proposed framework, summarize and compare these methods from different dimensions. 4) In Section~\ref{sec:imprv_tech}, we provide some possible variants within our framework to further improve the performance of existing methods. 5) In Section~\ref{sec:extension}, we present one possible implementation method for modeling inter-class correlation between representers. 6) In Section~\ref{sec:exp}, we validate the effectiveness of our method by several experiments. 7) In Section~\ref{sec:conclusion}, we conclude this work, analyze the limitations, and provide several further research directions.

\section{Preliminaries}
\label{sec:pre}
\subsection{Contrastive Vision-Language Models}
Contrastive vision-language models (CVLMs), which serve as a foundational class of vision-language models, align the representation spaces of image and text (language) through cross-modal contrastive learning. CLIP (contrastive language-image pre-training)~\cite{CLIP} is a typical representative of this type of method. Generally speaking, CVLMs such as CLIP employ a dual-stream network architecture, depicted in Figure~\ref{fig:CLIP}, which consists of an image encoder and a text encoder. Given a batch of $n$ paired image-text data, the respective image and text features are extracted by the image encoder and the text encoder, denoted as $F_i\in\mathbb{R}^{d}$ and $T_i\in\mathbb{R}^{d}$ for $i=1,\cdots,n$, where $d$ represents the feature dimension. The following contrastive loss function $\mathcal{L}$ is optimized to learn the parameters in image and text encoder:
\begin{align}
\mathcal{L}&=\mathcal{L}_\text{i2t} + \mathcal{L}_\text{t2i},\\
\mathcal{L}_\text{i2t}&=-\frac{1}{n}\sum\limits_{i=1}^n \log \frac{\exp(\text{cos}(F_i, T_i)/\tau)}{\sum\nolimits_{j=1}^n\exp(\text{cos}(F_i, T_j)/\tau)},\\
\mathcal{L}_\text{t2i}&=-\frac{1}{n}\sum\limits_{i=1}^n \log \frac{\exp(\text{cos}(F_i, T_i)/\tau)}{\sum\nolimits_{j=1}^n\exp(\text{cos}(F_j, T_i)/\tau)},
\end{align}
where $\tau$ is a learnable temperature parameter and $\text{cos}(\cdot,\cdot)$ is the cosine similarity function. The term $\mathcal{L}_\text{i2t}$ is to predict the corresponding paired text id given the image features, while the term $\mathcal{L}_\text{t2i}$ is to predict the corresponding paired image id given the text features.

\subsection{Problem Definition}
The problem of parameter-efficient adaptation of pre-trained CVLMs for low-shot image recognition task is defined as follows:
\begin{definition}[PEFT of CVLMs for Low-shot Classification]
	For a low-shot image recognition task, an image classification training set $\mathcal{D}_\text{train}$ and a testing set $\mathcal{D}_\text{test}$ are provided, where $\mathcal{D}_\text{train}$ only contains a few training samples. This classification problem is addressed by adapting a pre-trained vision-language model $\mathcal{M}$ on $\mathcal{D}_\text{train}$ in a parameter-efficient manner. Formally, a few learnable parameters $\theta$ is introduced to $\mathcal{M}$ to generate a classifier $f(\cdot, \mathcal{M}, \theta)$. The parameters are tuned on $\mathcal{D}_\text{train}$ by minimizing the following regularized empirical risk $\mathcal{L}(\theta)$, i.e., 
	\begin{equation}
	\min\limits_{\theta} \mathcal{L}(\theta)=L\big(f(\cdot, \mathcal{M}, \theta), \mathcal{D}_\text{train}\big) + \lambda\cdot R(\theta),\label{eq:rer}
	\end{equation}
	where $L$ denotes the empirical risk defined on $\mathcal{D}_\text{train}$ and $R$ denotes the regularization term with $\lambda$ the regularization weight. To be more general, we allow $|\mathcal{D}_\text{train}|=0$ or $\theta=\varnothing$. In these cases, Eq.~\eqref{eq:rer} is not necessary, as the classifier $f$ can be defined directly without the need of adaptation. Finally, the learned classifier $f$ is evaluated on the testing set $\mathcal{D}_\text{test}$.
\end{definition}

According to the setting of $\mathcal{D}_\text{train}$, the above task can be classified further. Generally, the training set is consisted of $N_1=CM$ labeled images and $N_2$ unlabeled images, with $C$ the class number and $K$ the number of labeled images per class. When $N_2=0, K=0$, the task is a \textbf{zero-shot recognition task}~\cite{CLIP,abs-2401-02460,Zhang0L24}; when $N_2=0, K>0$ and $K$ is small, the task is a \textbf{supervised few-shot recognition task}~\cite{CoOp,TipAdapter,APE}; when $N_2>0, K>0$ and $K$ is small, the task is a \textbf{semi-supervised few-shot recognition task}; when $N_2>0, K=0$, the task is an \textbf{unsupervised recognition task}~\cite{abs-2204-03649}. 

\begin{figure*}[thb]
	\centering
	\includegraphics[width=0.95\linewidth]{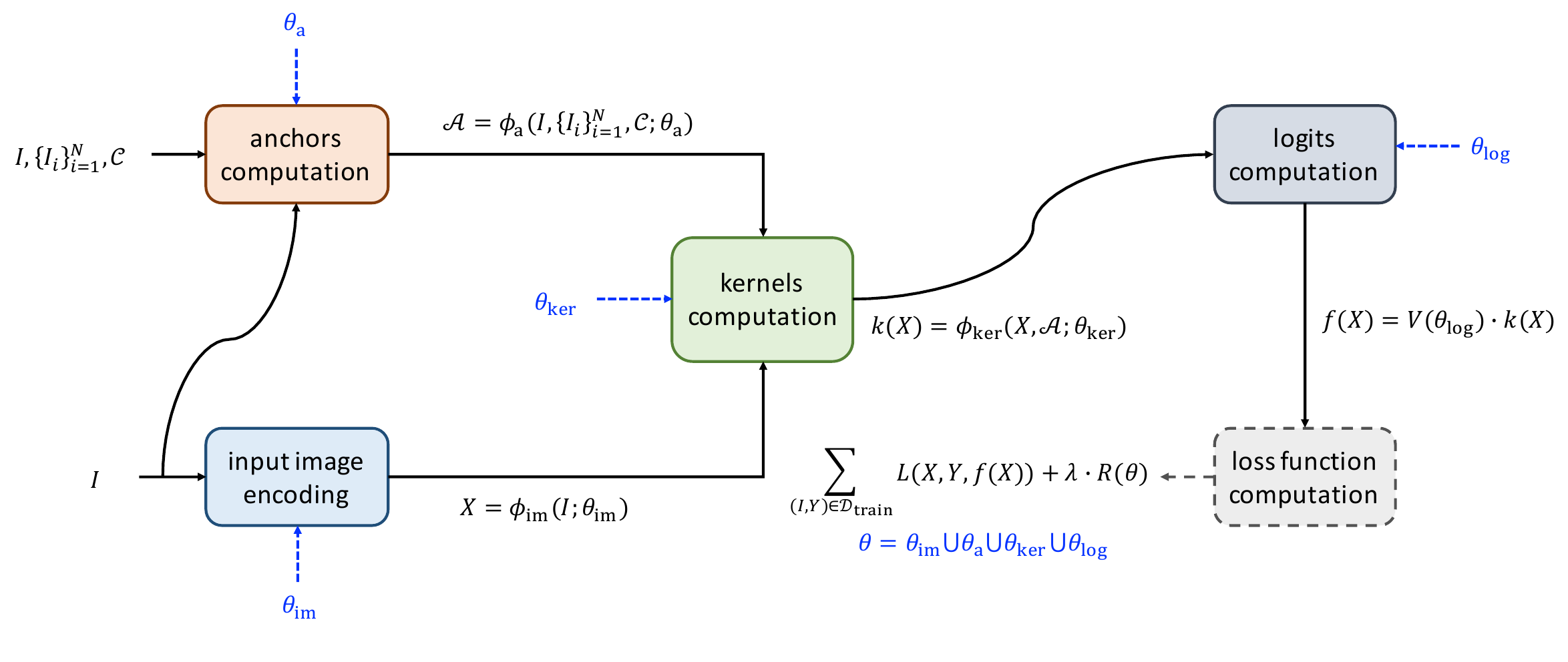}
	\caption{The proposed framework. For training-free methods, the loss function is not computed. $\theta_\text{a}, \theta_\text{im}, \theta_\text{ker}, \theta_\text{log}$ denote the learnable parameters in acnhors computation, input image encoding, kernels computation and logits computation, respectively. Other notations: $I$ the input image, $\{I_i\}_{i=1}^N$ the labeled and unlabeled training images, $\mathcal{C}$ the class names, $\mathcal{A}$ the anchors, $X$ the input image's features, $k(X)$ the kernel vector of $X$, $V(\theta_\text{log})$ the transformation matrix, $L$ the loss function, $R(\theta)$ the regularization term, $\lambda$ the regularization weight and $Y$ the ground-truth label.}
	\label{fig:framework}
\end{figure*}

\section{Unified Framework}
\label{sec:framework}

\subsection{The Representer Theorem}
For a learning problem, we consider an input space $\mathcal{X}$, an output space $\mathcal{Y}$ and $n$ training data $(X_1,Y_1),\cdots, (X_n, Y_n)$, where $X_i\in\mathcal{X}, Y_i\in\mathcal{Y}$. For regression problem, $\mathcal{Y}=\mathbb{R}$, and for binary classification problem $\mathcal{Y}=\{0,1\}$. We want to estimate a function $f:\mathcal{X}\rightarrow\mathcal{Y}$ to model the relation between input and output. Kernel method achieves this by minimizing the regularized empirical risk~\cite{ScholkopfHS01,DinuzzoS12}:
\begin{equation}
f^*=\arg\min_{f\in\mathcal{H}} \frac{1}{n}\sum\limits_{i=1}^n L\big(Y_i, f(X_i)\big) + \lambda \|f\|_\mathcal{H}^2,
\label{eq:reg}
\end{equation}
where $\mathcal{H}$ is a reproducing kernel Hilbert space (RKHS) associated to the kernel $k:\mathcal{X}\times \mathcal{X}\rightarrow \mathbb{R}$, $L$ is a loss function defined on each sample, $\|f\|_\mathcal{H}^2$ is a norm defined on RKHS that measures the complexity of $f$, and $\lambda$ is a penalty parameter. As $\mathcal{H}$ can be infinite dimensional, directly solving the above problem is infeasible. Thanks to the Representer Theorem~\cite{ScholkopfHS01,DinuzzoS12}, the above problem can be reduced to a finite dimensional optimization problem.
\begin{theorem}[The Representer Theorem]
	Given $n$ training samples $(X_1,Y_1),\cdots, (X_n, Y_n)$, consider the optimization problem in RKHS $\mathcal{H}$ with associated kernel $k$:
	\begin{align}
	\min\limits_{f\in\mathcal{H}} L\big(\{(X_i, Y_i, f(X_i))\}_{i=1}^n\big) + \psi(\|f\|_\mathcal{H}^2),
	\end{align}
	where $L$ is a loss function depends on $n$ examples, $\psi:[0,\infty)\rightarrow\mathbb{R}$ is a strictly increasing function. All minimizers $f^*$ of this problem admit the following form:
	\begin{align}
	f^*(X)=\sum\limits_{i=1}^n \alpha_i k(X_i, X), \alpha_i\in\mathbb{R}.
	\label{eq:rt_solution}
	\end{align}
\end{theorem}
The Representer Theorem states that the solution $f$ can be expressed by a linear combination of $n$ representers $k(X_i, \cdot)$, which are functions embedded in RKHS $\mathcal{H}$. This makes the intractable learning problem in RKHS become tractable. In fact, we only need to solve the new optimization problem about $\{\alpha_j\}_{j=1}^n$:
\begin{align}
\min\limits_{\{\alpha_j\}_{j=1}^n} &L\Big(\big\{(X_i, Y_i, \sum\limits_{j=1}^n \alpha_j k(X_i, X_j))\big\}_{i=1}^n\Big) \nonumber \\
&+ \psi\Big(\sum\limits_{i=1}^n\sum\limits_{j=1}^n\alpha_i\alpha_j k(X_i, X_j)\Big).
\label{eq:dual_prob}
\end{align}
Here, the second term is derived from the reproducing property of RKHS. Besides the above basic form of Representer Theorem, there are some further extensions in the literature. For example, Li~\cite{Lin1998} provided a semi-parametric extension for multicategory classification. Y. Lee \etal~\cite{MCSVM} presented the Representer Theorem for multicategory support vector machines. 

\subsection{Kernel Ridge Regression}
The kernel ridge regression (KRR) problem is a special case of Eq.~\eqref{eq:dual_prob} by letting $L(\{X_i,Y_i,f(X_i)\}_{i=1}^n)=\sum\nolimits_{i=1}^n (Y_i-f(X_i))^2$ and $\psi(t)=\lambda t$ with $\lambda$ a hyper-parameter. By some derivations, we obtain the following optimization problem:
\begin{align}
\min\limits_{\alpha} \alpha^T(K+\lambda \mathbb{I})\alpha -2Y^TK\alpha + Y^TY,
\end{align}
where $\alpha=[\alpha_1,\cdots, \alpha_{n}]^T$ collects all the coefficients, $Y=[Y_1,\cdots,Y_n]^T$ collects all the targets, $K=[k(X_i,X_j)]_{i,j=1}^n\in\mathbb{R}^{n\times n}$ is the kernel matrix, $\mathbb{I}$ is an identity matrix. The closed-form solution of KRR is
\begin{align}
\alpha^*=(K+\lambda \mathbb{I})^{-1}Y.
\end{align}
The prediction function of KRR is
\begin{align}
f(X)=(\alpha^*)^Tk(X)=Y^T(K+\lambda \mathbb{I})^{-1}k(X),
\end{align}
where $k(X)=[k(X, X_1),\cdots, k(X, X_n)]^T$ is the kernel vector. In this equation, $k(X)$ builds the relation between test sample and all training samples, $(K+\lambda \mathbb{I})^{-1}$ builds the relation between training samples. The solution can also be understood as an attention model, where $X$ is the query, $X_1,\cdots, X_n$ are the keys and $Y^T(K+\lambda \mathbb{I})^{-1}$ denote the values.

\subsection{From Representer Theorem to PEFT Framework}
In Eq.~\eqref{eq:rer}, the PEFT of CVLMs for low-shot classification targets at learning a function $f$ to generate the prediction for a new image, which is also a regularized empirical risk minimization problem. But different from kernel methods, we should apply the Representer Theorem to the extracted features by pre-trained CVLMs. Assume that $X$ and $X_1, \cdots, X_n$ are such image features and let us consider a $C$-way classification task, the optimal function $f$ for minimizing the associated empirical risk can be expressed in a similar form as Eq.~\eqref{eq:rt_solution}:
\begin{align}
f_j(X)=\sum\limits_{i=1}^n \alpha_{ij} k({X}_i, {X}),\label{eq:mulvar_rt}
\end{align}
where $f_j(X)$ is the $j$-th output and $\alpha_{ij}$ is coefficient, $j\in\{1,\cdots, C\}$. Note that the kernel $k({X}_i, {X})$ in the Representer Theorem is fixed, and the only learnable items are $\{\alpha_{ij}\}$. Such a constraint can limit the representation ability of the prediction function to a great extent. To solve this issue, we inject some learnable parameters while computing $X$ and $X_i$ by neural networks. To enable a parameter-efficient tuning, the number of the additionally introduced parameters should be controlled, either. 

To introduce the proposed framework, we now consider a $C$-way-$M$-shot classification task. The training dataset $\mathcal{D}_\text{train}$ consists of a labeled subset $\{(I_i, Y_i)\}_{i=1}^{N_1}$ ($N_1=CM$) and a unlabeled subset $\{I_i\}_{i=N_1+1}^{N}$, where $N=N_1+N_2$, $I_i$ is the $i$-th image and $Y_i\in\{1,\cdots,C\}$ is the corresponding label. For each class, there are $M$ labeled training images. Note that, when $N_2$ is zero, only labeled images are provided. 

The proposed unified PEFT framework of CVLMs for the low-shot image recognition task is represented by the following five equations:
\begin{align}
&\min\limits_{\theta} \mathcal{L}(\theta)=\sum\limits_{(I,Y)\in\mathcal{D}_\text{train}} {L}(X, Y, f(X)) + \lambda\cdot{R}(\theta),\label{eq:obj}\\
&\qquad\qquad\qquad f(X)={V}(\theta_\text{log})\cdot k(X), \label{eq:f}\\
&\!\!\!\!k(X)\!=\!\phi_\text{ker}(X,\mathcal{A};\theta_\text{ker})\!=\!\big[k_1(X,\!A_1),\!\cdots\!,\!k_n(X,\!A_n)\big]^T,\label{eq:k}\\
&\qquad\qquad\qquad\quad X=\phi_\text{im}(I; \theta_\text{im}),\label{eq:xi}\\
&\qquad\mathcal{A}=\{A_1, \cdots, A_n\} = \phi_\text{a}(I, \{I_i\}_{i=1}^N, \mathcal{C}; \theta_\text{a})\label{eq:A}.
\end{align}
Eq.~\eqref{eq:obj}-Eq.~\eqref{eq:A} mean loss function computation, logits computation, kernels computation, input image encoding, and anchors computation, respectively. Note that, when the training set $\mathcal{D}_\text{train}$ is empty, Eq.~\eqref{eq:obj} is not needed anymore.

\textbf{Loss Function Computation}. Eq.~\eqref{eq:obj} computes the regularized risk with $L$ the loss function defined on each training sample $(I, Y)$ and $R$ the regularization term defined on parameter $\theta$. $\lambda$ is a regularization weight. The symbol $\theta$ collects all learnable parameters in this framework, i.e., $\theta=\theta_\text{log}\cup\theta_\text{ker}\cup\theta_\text{im}\cup\theta_\text{a}$. For parameter-efficient adaptation, $\theta$ should only include a few parameters.

\textbf{Logits Computation}. Inspired by the Representer Theorem, we use a linear prediction function to generate the classification logits in Eq.~\eqref{eq:f}. By this way, $f$ lies in a subspace of RKHSes. ${V}(\theta_\text{log})$ is the transformation matrix, which has the same role to the $\alpha_{ij}$ in Eq.~\eqref{eq:mulvar_rt}. We assume that ${V}(\theta_\text{log})\in\mathbb{R}^{C\times n}$ is dependent on some learnable parameters collected by $\theta_\text{log}$, such that the prediction will be more flexible. In testing stage, the prediction label is determined by the index of the maximal value of $f(X)$.

\textbf{Kernels Computation}. In Eq.~\eqref{eq:k}, $k(X)\in\mathbb{R}^n$ collects all the representers in RKHS. As an extension, we do not constraint the kernel for each sample to be identical. In other words, each sample $A_j$ can have its own kernel $k_j$, enabling the usage of multiple kernels. To distinguish $A_j$ from normal training samples, we will refer to it as \textbf{anchor}. Note that, we also introduce learnable parameters collected by $\theta_\text{ker}$ to enhance the representation ability.

\textbf{Input image encoding}. Given an input image $I$, Eq.~\eqref{eq:xi} extracts its feature representation $X$ by a feature extractor $\phi_\text{im}$: $X=\phi_\text{im}(I; \theta_\text{im}) \in \mathbb{R}^{d\times n_\text{im}}$. Here, $\theta_\text{im}$ denotes the learnable parameters attached to $\phi_\text{im}$, $d$ is the dimension and $n_\text{im}$ is the number of image tokens. When pooling operation is adopted in $\phi_\text{im}$, $n_\text{im}$ is 1. Besides, in training stage, $I$ is from the training set, while in testing stage it is from the testing set.

\textbf{Anchors Computation}. Eq.~\eqref{eq:A} describes how to obtain a set of $n$ anchors for kernels computation. As the vision-language models are used, the anchors could include image anchors and text anchors. The image anchors are feature matrices in the output space of image encoder, while text anchors are feature matrices in the output space of text encoder. Let us assume that there are $n_1$ image anchors $\mathcal{A}_\text{im}=\{A_i^\text{im}\}_{i=1}^{n_1}$ with $A_i^\text{im}\in\mathbb{R}^{d\times n_\text{im}}$ the $i$-th image anchor, and there are $n_2$ text anchors $\mathcal{A}_\text{txt}=\{A_i^\text{txt}\}_{i=1}^{n_2}$ with ${A}_i^\text{txt}\in\mathbb{R}^{d\times n_\text{txt}}$ the $i$-th text anchor.  $n_\text{im}$ and $n_\text{txt}$ are the number of image and text tokens, respectively. Let $\mathcal{A}=\{A_i\}_{i=1}^{n}$ ($n=n_1+n_2$) denote the union of $\mathcal{A}_\text{im}$ and $\mathcal{A}_\text{txt}$. The anchors are computed by a function $\phi_\text{a}$, that is $\mathcal{A}=\phi_\text{a}(I, \{I_i\}_{i=1}^{N}, \mathcal{C}; \theta_\text{a})$, where $\mathcal{C}=\{\mathcal{C}_1,\cdots, \mathcal{C}_C\}$ collects all $C$ tokenized class names, $\theta_\text{a}$ collects all learnable parameters and $\{I_i\}_{i=1}^N$ denotes the set of all training images. We add $I$, the input image, as an extra input. By this way, the anchors can be made instance-conditioned. \textbf{Remark}: The anchors in classical kernel methods are obtained by random selection from training samples or by running clustering algorithm on these samples. In other words, the anchors and samples are from the same modality. Differently, we allow to use text anchors in our framework, which have different modality to the image features. The feasibility of using cross-modal anchors lies in the alignment of images and texts in the feature space achieved through pre-training.

The proposed framework is designed with a high level of abstraction, making it capable of integrating diverse approaches. Simultaneously, it incorporates the necessary granular details to facilitate meaningful comparisons between different approaches. The visual representation of this framework is depicted in Figure~\ref{fig:framework}.

\section{Instantiations}
\label{sec:instantiations}
Many of the existing zero/few-shot adaptation methods of CVLMs can be unified into the proposed framework. These include the prompt-based methods (CoOp~\cite{CoOp}, CoCoOp~\cite{CoCoOp}, KgCoOp~\cite{KgCoOp}, TaskRes~\cite{TaskRes}, MaPLe~\cite{MaPLe}, CALIP~\cite{CALIP}, LASP~\cite{LASP}) and adapter-based methods (CLIP-Adapter~\cite{CLIP_Adapter}, Tip-Adapter~\cite{TipAdapter}, Tip-Adapter-F~\cite{TipAdapter}, SuS-X~\cite{SuS}, CPR~\cite{CPR}, APE~\cite{APE}, APE-T~\cite{APE}). Note that, CALIP, Tip-Adapter, SUS-X and APE are training-free methods, while the rest are training-required methods. Besides, we also include two basic methods, LP-CLIP and zero-shot CLIP. The following sections demonstrate how to formulate these methods into the proposed framework by specializing each of the five components.

Following existing works, we use CLIP as the CVLM. Let us denote the frozen CLIP image encoder as $\text{E}_\text{v}()$ and frozen text encoder as $\text{E}_\text{t}()$. Both encoders output the embedding sequence of the last layer. We use $\text{P}()$ to denote the operation that extracts the pooled features from feature sequence. For image modality, the representation of the class token is used as pooled features, while for text modality the last text token's embedding is used.

\begin{figure*}
	\centering
	\subfloat[LP-CLIP]{\includegraphics[width=0.78\linewidth]{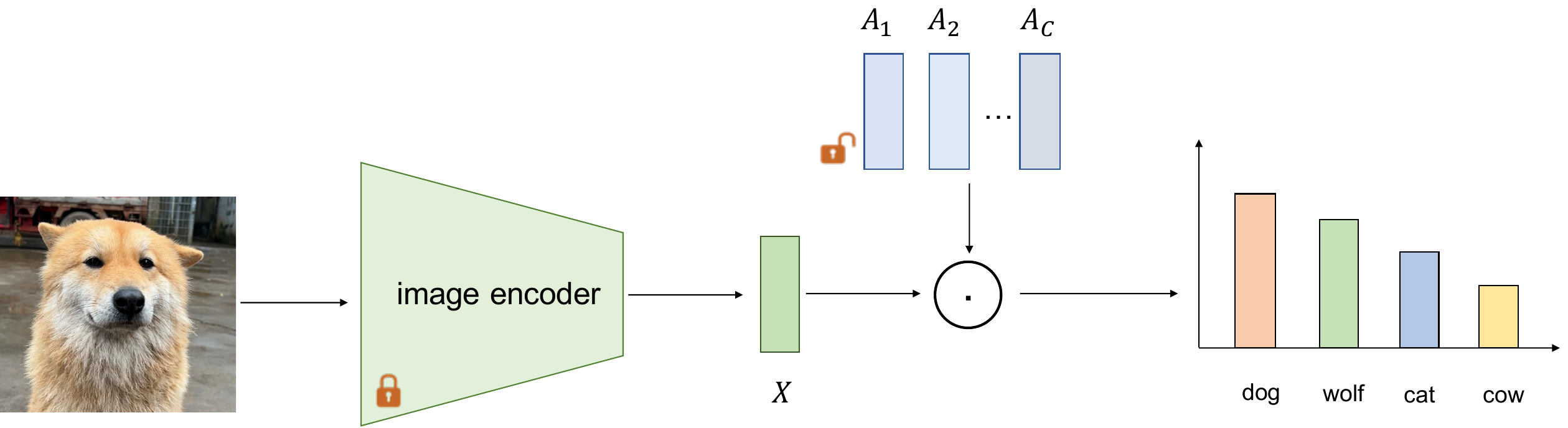}\label{fig:network_diff_methods_a}}
	\\
	\subfloat[Zero-shot CLIP]{\includegraphics[width=0.8\linewidth]{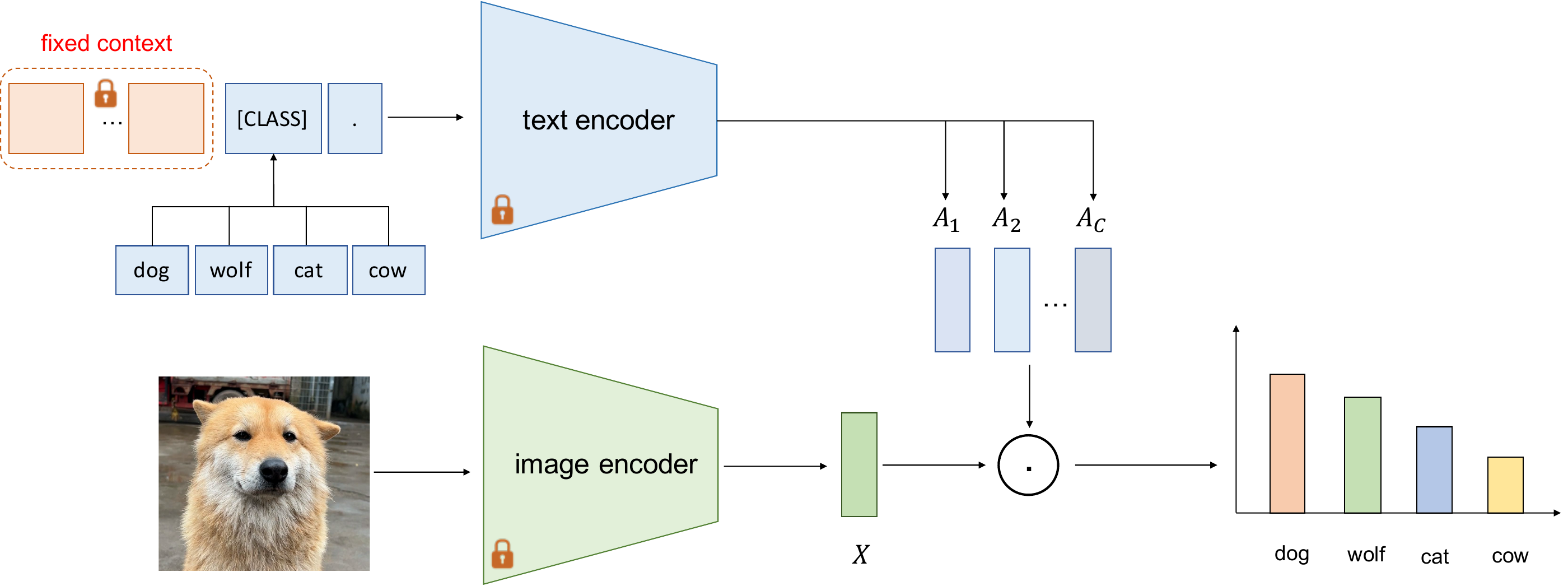}\label{fig:network_diff_methods_b}}
	\\
	\subfloat[CoOp]{\includegraphics[width=0.8\linewidth]{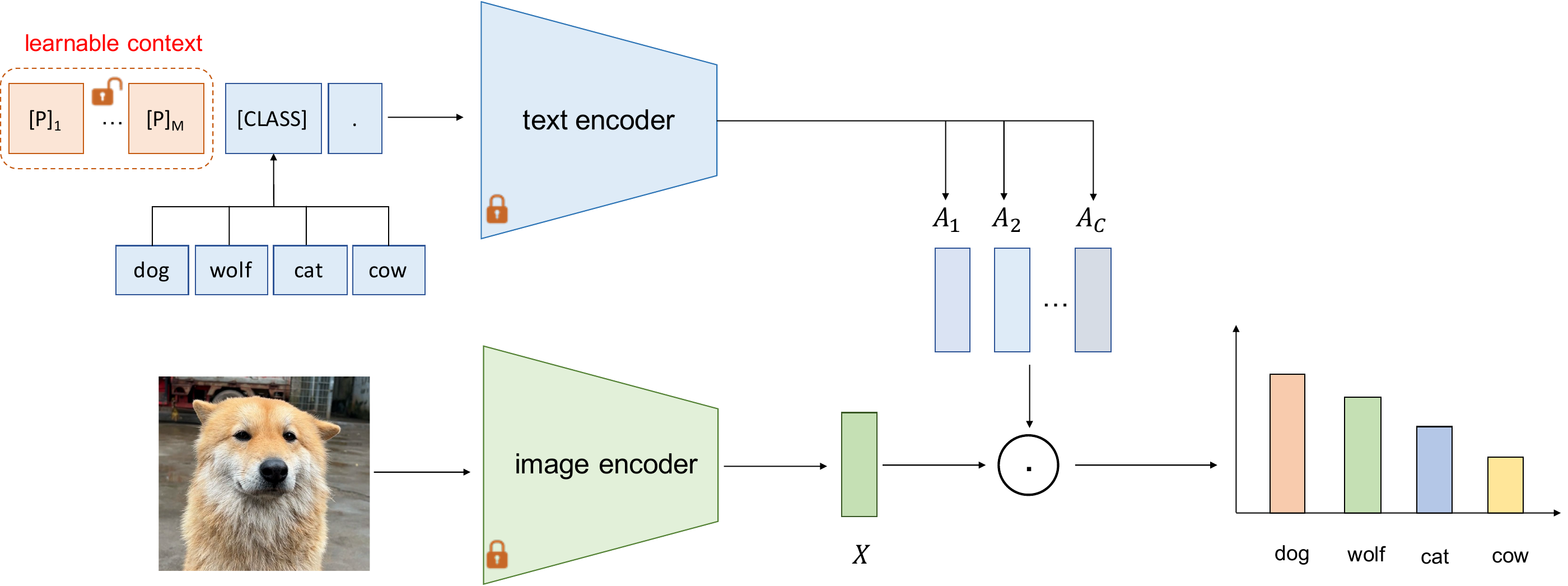}\label{fig:network_diff_methods_c}}
	\caption{Flowchart of LP-CLIP, Zero-shot CLIP and CoOp. In (a), $A_1,\cdots,A_C$ denote the learnable classifier weights; in (b) and (c), they are text features extracted by the text encoder, which also serve as the classifier weights. $X$ denotes the extracted image features by image encoder.}
	\label{fig:network_diff_methods}
\end{figure*}

\subsection{Basic Methods}
\subsubsection{LP-CLIP}
LP-CLIP~\cite{CLIP} is a strong baseline for few-shot classification exploiting pre-trained CLIP (ref. Figure~\ref{fig:network_diff_methods_a}), which performs linear classification with image features. Specifically, it extracts input image's features with frozen image encoder of CLIP, that is
\begin{align}
\phi_\text{im}(I; \theta_\text{im}) = \text{P}(\text{E}_\text{v}(I)),
\end{align}
where $\theta_\text{im}=\varnothing$. The image anchors are not used, thus, $n_1$ is zero. The $n_2$ text anchors $\mathcal{A}_\text{txt}=\{A_i^\text{txt}\}_{i=1}^{n_2}$ ($n_2=C$) are learned from scratch. Thus, we have $\theta_\text{a}=\mathcal{A}_\text{txt}$. The adopted kernel function is linear kernel, i.e.,
\begin{align}
k_i=k_\text{lin}, i=1,\cdots, n,
\end{align}
where $k_\text{lin}(X_1, X_2)=X_1^TX_2$ with $X_1, X_2\in\mathbb{R}^{d\times 1}$ and $n=n_1+n_2$. The kernels are not learnable, thus, we have $\theta_\text{ker}=\varnothing$. The transformation matrix to compute logits is identity matrix, that is
\begin{align}
V(\theta_\text{log})=\mathbb{I}\in \mathbb{R}^{C\times C},
\end{align}
where $\mathbb{I}$ is the identity matrix. The classification loss with weight decay is computed for training, thus we have 
\begin{align}
R(\theta)=\sum\limits_{c=1}^{n_2} \|{A}_c^\text{txt}\|_\text{F}^2,
\end{align}
where $\|\cdot\|_\text{F}$ denotes the Frobenius norm.  Besides, we have $\lambda>0$.

\subsubsection{Zero-shot CLIP}
Different from LP-CLIP, zero-shot CLIP~\cite{CLIP} (ref. Figure~\ref{fig:network_diff_methods_b}) extracts input image's L2-normalized features with frozen image encoder of CLIP:
\begin{align}
\phi_\text{im}(I; \theta_\text{im}) = \text{L2}\big(\text{P}(\text{E}_\text{v}(I))\big),
\end{align}
where $\text{L2}()$ is the columnwise L2-normalization operation. Thus, we have $\theta_\text{im}=\varnothing$. The image anchors are not used, thus, $n_1$ is zero. The text anchors are computed by
\begin{align}
A_c^\text{txt}=\text{L2}\big(\text{P}(\text{E}_\text{t}(\mathcal{C}_c))\big), c=1,\cdots,n_2,
\end{align}
where $n_2=C$. The parameter in $\phi_\text{a}$ is $\theta_\text{a}=\varnothing$. The adopted kernel function is linear kernel, i.e., $k_i=k_\text{lin}, i=1,\cdots,n$ with $n=n_1+n_2$. The kernels are not learnable, thus, we have $\theta_\text{ker}=\varnothing$. The transformation matrix to compute logits is identity matrix, that is $ V(\theta_\text{log})=\mathbb{I}\in \mathbb{R}^{C\times C}$. Zero-shot CLIP classifies an image according to the maximal value in the logits, thus, there is no need to compute loss function.

\subsection{Prompt-based Methods}

\subsubsection{CoOp}
Identical to zero-shot CLIP, CoOp~\cite{CoOp} (ref. Figure~\ref{fig:network_diff_methods_c}) uses CLIP's frozen image encoder to extract input image's features. CoOp only computes text anchors, thus we have $n_1=0$. The text anchors are obtained by frozen text encoder with learnable soft prompts of class names, i.e.,
\begin{align}
{A}_c^\text{txt}=\text{L2}\big(\text{P}(\text{E}_\text{t}([P,\mathcal{C}_c]))\big), c=1, \cdots, n_2,
\end{align}
where $n_2=C$, $P$ is a matrix containing several learnable soft prompt vectors and $[\cdot,\cdot]$ denotes horizontal concatenation of matrices. Therefore, we have $\theta_\text{a}=\{P\}$. The kernel function $k_i$ ($i=1,\cdots, n$ with $n=n_1+n_2$) is linear kernel, i.e., $k_i=k_\text{lin}$. Meantime, we have $\theta_\text{ker}=\varnothing$. The transformation matrix in CoOp is also identity matrix, i.e. $ V(\theta_\text{log})=\mathbb{I}$, which is fixed, thus $\theta_\text{log}=\varnothing$. CoOp uses the simple cross entropy loss without regularization term, as such, $\lambda$ is 0. The learnable parameter in CoOp can be represented as $\theta=\{P\}$.

\subsubsection{CoCoOp}
The only difference between CoCoOp~\cite{CoCoOp} and CoOp is that CoCoOp's soft prompts are conditioned on image features $X$. It uses a meta network $\text{Meta}()$ to learn bias of prompts and adds this bias to the original prompts $P$, that is the new prompt vectors are $P+\text{Meta}(X; \theta_\text{m})$ with $\theta_\text{m}$ the set of learnable parameters in the meta network. Therefore, we have
\begin{align}
A_c^\text{txt}=\text{L2}\big(\text{P}(\text{E}_\text{t}([P+\text{Meta}(X; \theta_\text{m}),\mathcal{C}_c]))\big), 
\end{align}
where $c=1,\cdots,n_2$ with $n_2=C$. As a result, the learnable parameter in CoCoOp can be represented as $\theta_\text{a}=\{P\}\cup \theta_\text{m}$.

\subsubsection{KgCoOp}
KgCoOp~\cite{KgCoOp} is developed based on CoOp and it introduces a regularization term on prompts to improve the generalization ability. The extra regularization term aims to reduce the gap between pre-trained text features of class names and text anchors, which is defined as
\begin{align}
R(\theta) = \sum\limits_{c=1}^{n_2} \|A_c^\text{txt}-{A}_c^\text{clip}\|_\text{F}^2,
\end{align}
where $A_c^\text{clip}=\text{L2}(\text{P}(\text{E}_\text{t}(\mathcal{C}_c))), c=1,\cdots,n_2$ with $n_2=C$ are the text features of class names extracted from pre-trained text encoder. Besides, the regularization term's weight meets $\lambda>0$. When $\lambda=0$, KgCoOp degrades to CoOp.

\subsubsection{TaskRes}
TaskRes~\cite{TaskRes} decouples the prior knowledge from pre-trained models and the new knowledge for a certain new task. For this aim, it adds a learnable task residual $R_c\in\mathbb{R}^{d\times 1}$ to the text features of class name $\mathcal{C}_c$ generated by pre-trained text encoder. Formally, the text anchor $A_c^\text{txt}$ is calculated by
\begin{align}
A_c^\text{txt}=\text{L2}\big(\text{P}(\text{E}_\text{t}(\mathcal{C}_c))+\gamma\cdot R_c\big), c=1,\cdots,n_2
\end{align}
where $\gamma$ is a scaling factor and $n_2=C$. Thus, the learnable parameter set of $\phi_\text{a}$ is $\theta_\text{a}=\{R_c\}_{c=1}^{n_2}$. The other settings of TaskRes are identical to CoOp.

\subsubsection{MaPLe}
MaPLe~\cite{MaPLe} is a multi-modal prompt tuning method. It does not have image anchors, thus, $n_1$ is 0. For the $i$-th layer ($i\in\{1,\cdots, l\}$) of text encoder, it introduces the input prompts ${P}_i$. Here, $l$ means that the first $l$ layers use prompts. The text anchors are computed by 
\begin{align}
A_c^\text{txt}=\text{L2}\big(\text{P}(\text{E}_\text{t}(\mathcal{C}_c;\{P_i\}_{i=1}^{l})\big),
\end{align}
where $c=1,\cdots, n_2$ with $n_2=C$. As a result, we have $\theta_\text{a}=\{P_i\}_{i=1}^{l}$. MaPLe transforms $P_i$ to generate the visual prompts of the image encoder's $i$-th layer, which is denoted by $\text{Trans}(P_i; \theta_i)$ with $\text{Trans}()$ the transforming function and $\theta_i$ denotes the set of learnable parameters for the $i$-th layer. As such, we have 
\begin{align}
\phi_\text{im}(I; \theta_\text{im}) = \text{L2}\big(\text{P}(\text{E}_\text{v}(I; \theta_\text{im}))\big), \theta_\text{im}=\cup_{i=1}^{l} \theta_{i}.
\end{align}
The transformation matrix $ V(\theta_\text{log})$ is identical to that in CoOp. The other settings of MaPLe are identical to CoOp.

\begin{figure*}
	\centering
	{\includegraphics[width=0.8\linewidth]{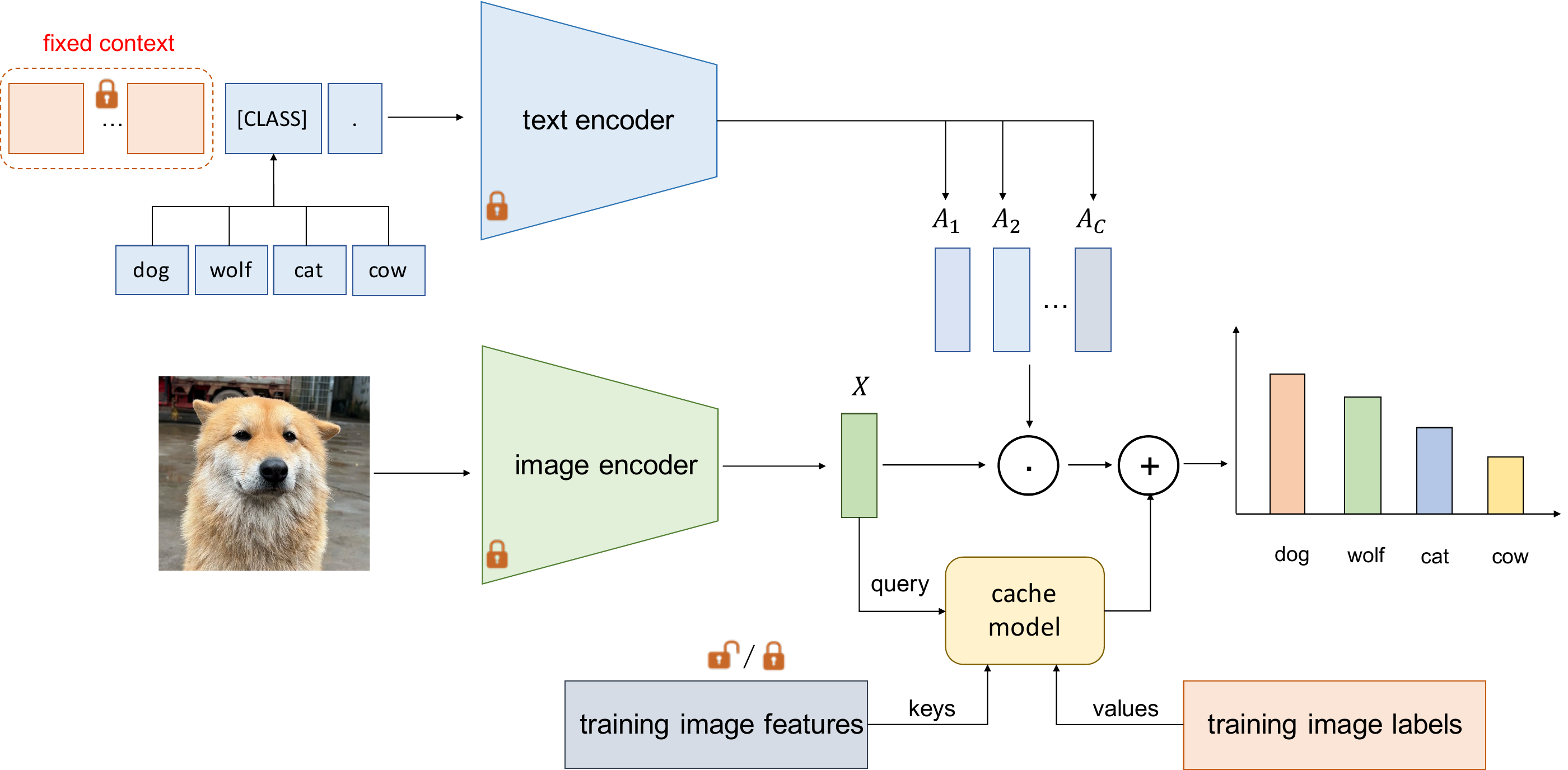}}
	\caption{Flowchart of Tip-Adapter and Tip-Adapter-F. For Tip-Adapter, the keys (i.e., the training image features) are frozen. For Tip-Adapter-F, the keys are learnable and initialized from the training image features.}
	\label{fig:network_diff_methods2}
\end{figure*}

\subsubsection{CALIP}
CALIP~\cite{CALIP} is zero-shot training-free adaptation method of CLIP, which does not need training images. It enhances the classification performance by introducing parameter-free cross-attention. Let us denote the visual and text features generated by the image and text encoder as ${F}_\text{v}=\text{E}_{\text{v}}(I)$ and ${F}_\text{t}=[\text{P}(\text{E}_{\text{t}}(\mathcal{C}_1)), \cdots, \text{P}(\text{E}_{\text{t}}(\mathcal{C}_C))]$, respectively. CALIP introduces a cross-modal module to enhance these two features. Using pooled ${F}_\text{v}$ as query, and ${F}_\text{t}$ as key and value, the updated image features are denoted as
\begin{align}
\tilde{F}_\text{v}=\text{CA}(\text{P}(F_\text{v}), F_\text{t}),
\end{align}
where $\text{CA}$ is the parameter-free cross-attention module. Similarly, the updated text features for the $c$-th class are denoted as 
\begin{align}
\tilde{F}_\text{t}[:, c]=\text{CA}(F_\text{t}[:,c], F_\text{v}),
\end{align}
where $F_\text{t}[:, c]$ is the $c$-th column of $F_\text{t}$. The final image features $X$ can be represented as 
\begin{align}
X=\text{L2}\big([\text{P}(F_\text{v}), \text{P}(F_\text{v}), \text{P}(\tilde{F}_\text{v})]\big).
\end{align}
The text anchors can be represented as 
\begin{align}
A_c^\text{txt}=\text{L2}\big([{F}_\text{t}[:, c], \tilde{{F}}_\text{t}[:, c], {F}_\text{t}[:, c]]\big), c=1,\cdots, n_2,
\end{align}
where $n_2=C$. The image anchors in CALIP are not defined, thus $n_1$ is 0. The adopted kernel function can be defined as
\begin{align}
k_\text{wlin}(X, A_c^\text{txt})=\sum\limits_{i=1}^3 \beta_{i}X[:, i]^TA^\text{txt}_c[:, i], c=1,\cdots,n_2, 
\end{align}
where $\beta_1, \beta_2,\beta_3$ are the weights and $n_2=C$. The kernel can be seen as a weighted linear kernel, which is also a linear kernel in nature. Besides, the transformation matrix in logits computation is $ V(\theta_\text{log})=\mathbb{I}$ with $\theta_{\text{log}}=\varnothing$.

\subsubsection{LASP}
LASP~\cite{LASP} is a state-of-the-art prompt tuning method for adapting CLIP. It creatively introduces a text-text contrastive loss to reduce the overfitting on training samples. For input image encoding, it uses the image encoder with tunable LN (layer normalization) layers. Thus, we have
\begin{align}
\phi_\text{im}(I; \theta_\text{im}) = \text{L2}\big(\text{P}(\text{E}_\text{v}(I; \theta_\text{im}))\big), 
\end{align}
where $\theta_\text{im}$ collects the parameters in the LN layers of the image encoder. LASP only computes text anchors, thus $n_1=0$. Compared to CoOp, LASP introduces a learnable bias to the text features:
\begin{align}
A_c^\text{txt}=\text{L2}\big(\text{P}(\text{E}_\text{t}([P,\mathcal{C}_c]))+B\big), c=1,\cdots,n_2,
\end{align}
where $n_2=C$, $P$ contains several learnable soft prompt vectors and $B$ is a shared bias across all classes. Therefore, we have $\theta_\text{a}=\{P, B\}$. To compute the text-text contrastive loss, $Q$ hand-crafted text prompts are provided, i.e., $P_1,\cdots, P_Q$. The text features for the $c$-th class name with $P_q$ ($q\in\{1,\cdots,Q\}$) as its prefix prompts are computed by $A_{qc}^\text{clip}=\text{L2}(\text{P}(\text{E}_\text{t}([{P}_q,\mathcal{C}_c])))$. The text-text contrastive loss is defined as
\begin{align}
R(\theta)=-\frac{1}{Q}\sum\limits_{q=1}^Q \sum\limits_{c=1}^C\log \left(\frac{\exp((A^\text{txt}_c)^T A^\text{clip}_{qc}/\tau)}{\sum\nolimits_{i=1}^C \exp((A^\text{txt}_i)^T A^\text{clip}_{qi}/\tau)}\right),
\end{align}
where $\tau$ is a temperature parameter. Besides, the transformation matrix $ V(\theta_\text{log})$ is identical to that of CoOp.

\subsection{Adapter-based Methods}
\subsubsection{CLIP-Adapter}
CLIP-Adapter~\cite{CLIP_Adapter} (ref. Figure~\ref{fig:network_diff_methods2}) employs frozen image and text encoders, while updating the image and text features of the final layers through learnable adapter modules. For some image features $X$, the updated features are
\begin{align}
\text{Ada}_\text{v}(X;\{{W}_1^\text{v}, {W}_2^\text{v}\})&=\alpha_\text{v}\cdot {W}_2^\text{v} \cdot \text{ReLU}({W}_1^\text{v}X)\nonumber\\
&\quad+ (1-\alpha_\text{v})\cdot X, 
\end{align}
where ${W}_1^\text{v}$ and ${W}_2^\text{v}$ are learnable parameters, $\text{ReLU}$ is rectified linear unit and $\alpha_\text{v}$ is a hyper-parameter. Similarly, given text features $X$, the updated text features are
\begin{align}
\text{Ada}_\text{t}(X;\{{W}_1^\text{t}, {W}_2^\text{t}\})&=\alpha_\text{t}\cdot {W}_2^\text{t}\cdot\text{ReLU}({W}_1^\text{t}X)\nonumber\\
&\quad+ (1-\alpha_\text{t})\cdot X,
\end{align}
where ${W}_1^\text{t}$ and ${W}_2^\text{t}$ are learnable parameters, and $\alpha_\text{t}$ is a hyper-parameter. Finally, image features and text anchors of CLIP-Adapter are respectively computed by
\begin{align}
X&=\text{L2}\big(\text{Ada}_\text{v}(\text{P}(\text{E}_\text{v}(I)); \{{W}_1^\text{v}, {W}_2^\text{v}\})\big),\\
A^\text{txt}_c&=\text{L2}\big(\text{Ada}_\text{t}(\text{P}(\text{E}_\text{t}(\mathcal{C}_c)); \{{W}_1^\text{t}, {W}_2^\text{t}\})\big),
\end{align}
where $c\in\{1,\cdots,n_2\}$ with $n_2=C$. Thus, we have $\theta_\text{im}=\{{W}_1^\text{v}, {W}_2^\text{v}\}$ and $\theta_\text{a}=\{{W}_1^\text{t}, {W}_2^\text{t}\}$. Except the above changes, the other parts of CLIP-Adapter are identical to CoOp.

\subsubsection{Tip-Adapter and Tip-Adapter-F}
Tip-Adapter~\cite{TipAdapter} and Tip-Adapter-F~\cite{TipAdapter} are training-free and training-required adapter-based methods with key-value cache. The introduction of this cache amounts to defining kernel features for input image features based on image anchors. In Tip-Adapter, the image anchors (also the keys) are computed by pre-trained CLIP's image encoder, that is
\begin{align}
A^\text{im}_i=\text{L2}\big(\text{P}(\text{E}_\text{v}(I_i))\big), i=1, \cdots, n_1,\label{eq:keys}
\end{align}
where $n_1=CM$. The text anchors are computed by $A_c^\text{txt}=\text{L2}(\text{P}(\text{E}_\text{t}(\mathcal{C}_c))), c=1,\cdots,n_2$ with $n_2=C$. As such, the function $\phi_\text{a}$ has the parameter $\theta_\text{a}=\varnothing$. The kernel functions in Tip-Adapter are defined as
\begin{align}
k_i=\begin{cases}
k_\text{gau},&i\in\{1, \cdots,n_1\}\\
k_\text{lin},&i\in\{n_1,\cdots, n\}
\end{cases},
\end{align}
where $n=n_1+n_2$ and $k_\text{gau}$ is the Gaussian kernel function, which is defined as $k_\text{gau}=\exp(-\beta\|X_1-X_2\|^2_\text{F}/2)$. Note that, both $X_1$ and $X_2$ are L2-normalized features with the same shape. The kernels are not learnable, thus we have $\theta_\text{ker}=\varnothing$.

Tip-Adapter uses the one-hot label matrix $ Z\in\mathbb{R}^{C\times n_1}$ of training samples as the values of the cache model, which is equivalent to define the following transformation matrix in our framework:
\begin{align}
V(\theta_\text{log})=[\alpha Z,\mathbb{I}], 
\end{align}
where $\alpha$ is a hyper-parameter and $\theta_\text{log}$ is $\varnothing$. Finally, as Tip-Adapter is training-free, thus, $\theta=\varnothing$.

As for Tip-Adapter-F, the keys $\mathcal{A}_\text{im}=\{A_i^\text{im}\}_{i=1}^{n_1}$ (i.e., image anchors) are optimized directly, which are initialized by Eq.~\eqref{eq:keys}. Thus, we have $\theta=\theta_\text{a}=\mathcal{A}_\text{im}$.

\subsubsection{SuS-X}
SuS-X~\cite{SuS} is a training-free and name-only adaptation method of CLIP. Although it is developed based on Tip-Adapter, it does not require real training images in the training-free mode like Tip-Adapter. Thus, it can be viewed as a zero-shot adaptation method. The method to encode input image is the same as Tip-Adapter. SuS-X avoids using real training images by synthesizing them with a pre-trained stable diffusion model~\cite{StableDiffusion}, or by retrieving similar images with text-to-image retrieval over a large archive, i.e., LAION-5B~\cite{LAION5B}. Take the generation by diffusion model as an example, the generated $CM$ images are denoted as $I_{i}^\text{syn}, i=1,\cdots, CM$. Here, we assume the index range $(c-1)M+1,\cdots,cM$ corresponds to the $c$-th class. The $i$-th image anchor is computed by
\begin{align}
A_i^\text{im}=\text{L2}\big(\text{P}(\text{E}_\text{v}(I_i^\text{syn}))\big), i=1,\cdots,n_1,
\end{align}
where $n_1$ is $CM$. The $n_2=C$ text anchors are computed like Tip-Adapter. SuS-X introduces a special design to calibrate the kernel between image features and image anchors. To be specific, the calibrated kernel is defined as
\begin{align}
k_\text{calib-gau}(X_1, X_2)&=\alpha k_\text{gau}(X_1, X_2) \nonumber \\
&\quad+ \gamma\cdot\phi(-\text{KL}(X_1^T{W}|X_2^T{W})),
\end{align}
where $X_1, X_2\in \mathbb{R}^{d\times1}$, $\gamma$ is a hyper-parameter, $\phi$ is a re-scaling function, ${W}=[\text{L2}(\text{P}(\text{E}_\text{t}(\mathcal{C}_1))), \cdots, \text{L2}(\text{P}(\text{E}_\text{t}(\mathcal{C}_C)))]$. The final kernel functions in SuS-X are
\begin{align}
k_i=\begin{cases}
k_\text{calib-gau},&i\in\{1,\cdots,n_1\}\\
k_\text{lin},&i\in\{n_1, \cdots, n_1+n_2\},
\end{cases}
\end{align}
where $n_1=CM$ and $n_2=C$. Accordingly, the transformation matrix is $ V(\theta_\text{log})=[ Z,\mathbb{I}]$.

\subsubsection{CPR}
CPR~\cite{CPR} utilizes frozen image and text encoders but introduces learnable instance-conditioned residual to text features. For an input image $I$, the extracted image features are expressed as $X=\text{L2}(\text{P}(\text{E}_\text{v}(I, \theta_\text{im})))$ with $\theta_\text{im}=\varnothing$.

Given $n_2=C$ class names, their text features are $\{\text{P}(\text{E}_\text{t}(\mathcal{C}_c))\}_{c=1}^C$. For the $c$-th class, CPR introduces an instance-conditioned residual $B_c^\text{txt}$ generated by attending test image features to text features:
\begin{align}
B_c^\text{txt}=\text{CA}(\text{E}_\text{v}(I), \{\text{P}(\text{E}_\text{t}(\mathcal{C}_c))\}_{c=1}^C; \theta_1),
\end{align}
where $\text{CA}$ is a cross-attention block parameterized by $\theta_1$. The first parameter of $\text{CA}$ represents query, the second parameter is key and value. 
CPR also introduces an instance-conditioned residual $B_c^\text{im}$ generated by attending test image features to training image features:
\begin{align}
B_c^\text{im}=\text{CA}(\text{E}_\text{v}(I), \{\text{P}(\text{E}_\text{v}(I_c))\}_{c=1}^C; \theta_2),
\end{align}
where $\theta_2$ collects the learnable parameters and $I_c$ is a training image from the $c$-th class. Thus, the instance-conditioned text anchor for the $c$-th class is represented as
\begin{align}
A_c^\text{txt}&=\text{E}_\text{t}(\mathcal{C}_c) + B_c^\text{txt} + B_c^\text{im}.
\end{align}

CPR uses features of unlabeled images to construct the image anchors. The features of the $j$-th ($j\in\{N_1, \cdots, N_1+N_2\}$) image are $\text{E}_\text{v}(I_j)$. CPR uses $A_c^\text{txt}$ to retrieve $m$ nearest neighbors $\text{E}_\text{v}(I_j), j\in \mathcal{I}_c\subseteq \{N_1, \cdots, N_1+N_2\}, |\mathcal{I}_c|=m$. The image anchor for the $c$-th class ($c=1,\cdots,n_1$ with $n_1=C$) is defined as
\begin{align}
A^\text{im}_c=\frac{1}{m}\sum\limits_{t\in \mathcal{I}_c}\text{E}_\text{v}(I_t),
\end{align}
Thus, the function $\phi_\text{a}$ has the parameter $\theta_\text{a}=\theta_1\cup\theta_2$.

CPR introduces a hyper-parameter $\alpha$ to balance $A^\text{txt}_c$ and $A^\text{im}_c$, that is $A_c'=\alpha A^\text{txt}_c+(1-\alpha)A_c^\text{im}$. The logits in CPR can be represented as
\begin{align}
f(X)&=\left[\frac{X^TA_1'}{\|X\|_\text{F}\|A_1'\|_\text{F}}, \cdots, \frac{X^TA_C'}{\|X\|_\text{F}\|A_C'\|_\text{F}}\right]^T\nonumber\\
&={\Gamma}[X^TA_1',\cdots, X^TA_C']^T\nonumber\\
&={\Gamma}\left\{\alpha[X^TA_1^\text{txt},\cdots, X^TA_C^\text{txt}]^T\right.\nonumber\\
&\quad+\left.(1-\alpha)[X^TA_1^\text{im},\cdots, X^TA_C^\text{im}]^T\right\}\nonumber\\
&=[(1-\alpha){\Gamma},\alpha{\Gamma}]\begin{bmatrix}k_\text{im}(X)\\k_\text{txt}(X)\end{bmatrix}\nonumber\\
&= V(\theta_\text{log})\cdot k(X)
\end{align}
where $V(\theta_\text{log})=[(1-\alpha){\Gamma},\alpha{\Gamma}]$ and 
\begin{align}
\Gamma&=\text{diag}\Big(\big[\frac{1}{\|X\|_\text{F}\|A_1'\|_\text{F}},\cdots, \frac{1}{\|X\|_\text{F}\|A_C'\|_\text{F}}\big]\Big),\\
k_\text{im}(X)&=[X^TA_1^\text{im},\cdots,X^TA_c^\text{im},\cdots, X^TA_C^\text{im}]^T,\\
k_\text{txt}(X)&=[X^TA_1^\text{txt},\cdots,X^TA_c^\text{txt},\cdots, X^TA_C^\text{txt}]^T.
\end{align}
Obviously, the kernels in CPR are $k_i=k_\text{lin}, i=1,\cdots,n$ with $n=n_1+n_2$. It is worth nothing that $\Gamma$ and $ V(\theta_\text{log})$ are dependent on $X$. The kernels are not learnable, thus, we have $\theta_\text{ker}=\varnothing$. The transformation matrix is also fixed, thus, $\theta_\text{log}=\varnothing$.

As for model training, CPR also introduces a KgCoOp-style regularization term on parameters. However, $\{A_c^\text{clip}\}_{c=1}^C$ are computed through ensembling many prompt contexts, which are generated by a pre-trained LLM. The learnable parameters in CPR are denoted as $\theta=\theta_1\cup\theta_2$.

\begin{table*}
	\caption{Comparison of different adaptation methods of CLIP. Blue color marks the differences to zero-shot CLIP. \colorbox{LightRed}{\color{LightRed}{X}}: input image encoding without learnable parameters; \colorbox{LightOrange}{\color{LightOrange}{X}}: input image encoding with learnable parameters; \colorbox{LightYellow}{\color{LightYellow}{X}}: only use text anchors; \colorbox{LightGreen}{\color{LightGreen}{X}}: use text and image anchors simultaneously; \colorbox{LightCyan}{\color{LightCyan}{X}}: only use linear kernel; \colorbox{LightBlue}{\color{LightBlue}{X}}: use non-linear kernel; \colorbox{LightIndigo}{\color{LightIndigo}{X}}: non-learnable transformation matrix; \colorbox{LightMagenta}{\color{LightMagenta}{X}}: learnable transformation matrix; \colorbox{babypink}{\color{babypink}{X}}: training-based methods with regularization; \colorbox{almond}{\color{almond}{X}}: training-free methods; \colorbox{blond}{\color{blond}{X}}: training-based methods without regularization.}
	\centering
	\fontsize{7}{8.4}\selectfont
	\renewcommand{\arraystretch}{1.5}
	\addtolength{\tabcolsep}{-0.7em}
	\begin{tabular}{|p{15mm}|p{32mm}|p{54mm}|p{25mm}|p{34mm}|p{12mm}|}
		\hline
		\makecell{\textbf{Method}} & \makecell{\textbf{input image}\\\textbf{encoding} $\phi_\text{im}$} & \makecell{\textbf{anchors} \\$\mathcal{A}=\mathcal{A}_\text{im}\cup\mathcal{A}_\text{txt}$} & \makecell{\textbf{kernels}} & \makecell{\textbf{logits}\\$f(X)\!=\!V(\theta_\text{log})k(X)$}& \makecell{$\mathcal{L}(\theta)$}\\
		\hline
		\makecell{LP-CLIP\\\cite{CLIP}} & \cellcolor{LightRed} \makecell{$\text{L2}(\text{E}_\text{v}(I))$\\$\theta_\text{im}=\varnothing$}& \cellcolor{LightYellow} \makecell{$n_1=0, n_2=C$\\$\color{blue}{\theta_\text{a}=\mathcal{A}_\text{txt}}$} & \cellcolor{LightCyan} \makecell{$k_i\!=\!k_\text{lin}$\\$i=1,\cdots,n_1\!+\!n_2$\\${\theta_\text{ker}=\varnothing}$} & \cellcolor{LightIndigo} \makecell{$ V(\theta_\text{log})\!=\!\mathbb{I}$\\$\theta_\text{log}=\varnothing$} &\cellcolor{babypink} \makecell{$\color{blue}{\lambda>0}$}\\
		\hline
		\makecell{Zero-shot\\CLIP\cite{CLIP}} & \cellcolor{LightRed} \makecell{$\text{L2}(\text{P}(\text{E}_\text{v}(I)))$\\$\theta_\text{im}=\varnothing$}&\cellcolor{LightYellow} \makecell{$n_1=0,n_2=C$\\$A_c^\text{txt}\!=\!{\text{L2}(\text{P}(\text{E}_\text{t}(\mathcal{C}_c)))}$\\${\theta_\text{a}=\varnothing}$} &\cellcolor{LightCyan} \makecell{$k_i\!=\!k_\text{lin}$\\$i=1,\cdots,n_1\!+\!n_2$\\${\theta_\text{ker}=\varnothing}$} &  \cellcolor{LightIndigo} \makecell{$ V(\theta_\text{log})\!=\!\mathbb{I}$\\$\theta_\text{log}=\varnothing$} &\cellcolor{almond}\makecell{N/A}\\
		\hline
		\hline
		\makecell{CoOp\cite{CoOp} }& \cellcolor{LightRed} \makecell{$\text{L2}(\text{P}(\text{E}_\text{v}(I)))$\\$\theta_\text{im}=\varnothing$}&\cellcolor{LightYellow} \makecell{$n_1=0,n_2=C$\\$\color{blue}{A_c^\text{txt}=\text{L2}(\text{P}(\text{E}_\text{t}([P,\mathcal{C}_c])))}$\\$\color{blue}{\theta_\text{a}=\{P\}}$} &\cellcolor{LightCyan} \makecell{$k_i\!=\!k_\text{lin}$\\$i=1,\cdots,n_1\!+\!n_2$\\${\theta_\text{ker}=\varnothing}$} &  \cellcolor{LightIndigo} \makecell{$ V(\theta_\text{log})\!=\!\mathbb{I}$\\$\theta_\text{log}=\varnothing$} &\cellcolor{blond} \makecell{$\color{blue}{\lambda=0}$}\\
		\hline
		\makecell{CoCoOp\\\cite{CoCoOp} }&\cellcolor{LightRed} \makecell{$\text{L2}(\text{P}(\text{E}_\text{v}(I)))$\\$\theta_\text{im}=\varnothing$} &\cellcolor{LightYellow} \makecell{$n_1=0,n_2=C$\\$\color{blue}{A_c^\text{txt}\!=\!\text{L2}(\text{P}(\text{E}_\text{t}([P\!+\!\text{Meta}(X; \theta_\text{m}),\mathcal{C}_c])))}$\\$\color{blue}{\theta_\text{a}\!=\!\{P\}\cup\theta_\text{m}}$} &\cellcolor{LightCyan} \makecell{$k_i=k_\text{lin}$\\$i=1,\cdots,n_1\!+\!n_2$\\${\theta_\text{ker}=\varnothing}$} &  \cellcolor{LightIndigo} \makecell{$ V(\theta_\text{log})\!=\!\mathbb{I}$\\$\theta_\text{log}=\varnothing$} & \cellcolor{blond}\makecell{$\color{blue}{\lambda=0}$}\\
		\hline
		\makecell{KgCoOp\\\cite{KgCoOp}} &\cellcolor{LightRed} \makecell{$\text{L2}(\text{P}(\text{E}_\text{v}(I)))$\\$\theta_\text{im}=\varnothing$}&\cellcolor{LightYellow} \makecell{$n_1=0,n_2=C$\\$\color{blue}{A_c^\text{txt}=\text{L2}(\text{P}(\text{E}_\text{t}([P,\mathcal{C}_c])))}$\\$\color{blue}{\theta_\text{a}=\{P\}}$} &\cellcolor{LightCyan} \makecell{$k_i=k_\text{lin}$\\$i=1,\cdots,n_1\!+\!n_2$\\${\theta_\text{ker}=\varnothing}$} & \cellcolor{LightIndigo}  \makecell{$ V(\theta_\text{log})\!=\!\mathbb{I}$\\$\theta_\text{log}=\varnothing$} & \cellcolor{babypink}\makecell{$\color{blue}{\lambda>0}$}\\
		\hline
		\makecell{TaskRes\\\cite{TaskRes} }&\cellcolor{LightRed} \makecell{$\text{L2}(\text{P}(\text{E}_\text{v}(I)))$\\$\theta_\text{im}=\varnothing$} &\cellcolor{LightYellow} \makecell{$n_1=0,n_2=C$\\$\color{blue}{A_c^\text{txt}=\text{L2}(\text{P}(\text{E}_\text{t}(\mathcal{C}_c))+\gamma\cdot R_c)}$\\$\color{blue}{\theta_\text{a}\!=\!\{R_c\}_{c=1}^{n_2}}$} & \cellcolor{LightCyan}\makecell{$k_i=k_\text{lin}$\\$i=1,\cdots, n_1\!+\!n_2$\\${\theta_\text{ker}=\varnothing}$} & \cellcolor{LightIndigo}  \makecell{$ V(\theta_\text{log})\!=\!\mathbb{I}$\\$\theta_\text{log}=\varnothing$} &\cellcolor{blond} \makecell{$\color{blue}{\lambda=0}$}\\
		\hline
		\makecell{MaPLe\\\cite{MaPLe}}& \cellcolor{LightOrange} \makecell{$\color{blue}{\text{L2}(\text{P}(\text{E}_\text{v}(I; \theta_\text{im})))}$\\$\color{blue}{\theta_\text{im}=\bigcup_{i=1}^{l} \theta_{i}}$}  &\cellcolor{LightYellow} \makecell{$n_1=0,n_2=C$\\$\color{blue}{A_c^\text{txt}=\text{L2}(\text{P}(\text{E}_\text{t}(\mathcal{C}_c; \{P_1,\cdots, P_{l}\})))}$\\$\color{blue}{\theta_\text{a}\!=\!\{P_1,\cdots,P_{l}\}}$} & \cellcolor{LightCyan}\makecell{$k_i=k_\text{lin}$\\$i=1,\cdots,n_1\!+\!n_2$\\${\theta_\text{ker}=\varnothing}$} &  \cellcolor{LightIndigo} \makecell{$ V(\theta_\text{log})\!=\!\mathbb{I}$\\$\theta_\text{log}=\varnothing$} & \cellcolor{blond}\makecell{$\color{blue}{\lambda=0}$} \\
		\hline
		\makecell{CALIP\\\cite{CALIP}} & \cellcolor{LightOrange} \makecell{\color{blue}{$\text{L2}([\text{P}(F_v),\!\text{P}(F_v),\!\text{P}(\tilde{F}_v)])$}\\\color{blue}{$\theta_\text{im}=\varnothing$}} &\cellcolor{LightYellow} \makecell{{$n_1=0,n_2=C$}\\\color{blue}{$A_c^\text{txt}\!=\!\text{L2}([F_t[:, c], \tilde{F}_t[:, c], F_t[:, c]])$}\\\color{blue}{$\theta_\text{a}=\varnothing$}} &\cellcolor{LightCyan} \makecell{\color{blue}{$k_i=k_\text{wlin}$}\\$i=1,\cdots,n_1\!+\!n_2$\\${\theta_\text{ker}=\varnothing}$} &  \cellcolor{LightIndigo} \makecell{$ V(\theta_\text{log})\!=\!\mathbb{I}$\\$\theta_\text{log}=\varnothing$} &\cellcolor{almond} \makecell{N/A}\\
		\hline
		\makecell{LASP\\\cite{CoOp} }&  \cellcolor{LightOrange} \makecell{$\text{L2}(\text{P}(\text{E}_\text{v}(I; \theta_\text{im})))$\\$\color{blue}{\theta_\text{im}\neq\varnothing}$}&\cellcolor{LightYellow} \makecell{$n_1=0,n_2=C$\\$\color{blue}{A_c^\text{txt}=\text{L2}(\text{P}(\text{E}_\text{t}([P,\mathcal{C}_c]))+B)}$\\$\color{blue}{\theta_\text{a}=\{P,B\}}$} &\cellcolor{LightCyan} \makecell{$k_i\!=\!k_\text{lin}$\\$i=1,\cdots,n_1\!+\!n_2$\\${\theta_\text{ker}=\varnothing}$} &  \cellcolor{LightIndigo} \makecell{$ V(\theta_\text{log})\!=\!\mathbb{I}$\\$\theta_\text{log}=\varnothing$} & \cellcolor{babypink}\makecell{$\color{blue}{\lambda>0}$}\\
		\hline
		\hline
		\makecell{\makecell{CLIP-\\Adapter\\\cite{CLIP_Adapter}}}& \cellcolor{LightOrange}  \makecell{$\color{blue}{\text{L2}(\text{Ada}_\text{v}(\text{P}(\text{E}_\text{v}(I));\theta_\text{im})}$\\$\color{blue}{\theta_\text{im}\!=\!\{W_1^\text{v},\!W_2^\text{v}\}}$} &\cellcolor{LightYellow} \makecell{$n_1=0,n_2=C$\\$\color{blue}{A_c^\text{txt}=\text{L2}(\text{Ada}_\text{t}(\text{P}(\text{E}_\text{t}(\mathcal{C}_c));\theta_\text{a})}$\\$\color{blue}{\theta_\text{a}\!=\!\{W_1^\text{t}, W_2^\text{t}\}}$} & \cellcolor{LightCyan}\makecell{$k_i\!=\!k_\text{lin}$\\$i=1,\cdots,n_1\!+\!n_2$\\${\theta_\text{ker}=\varnothing}$} &  \cellcolor{LightIndigo} \makecell{$ V(\theta_\text{log})\!=\!\mathbb{I}$\\$\theta_\text{log}=\varnothing$} &\cellcolor{blond} \makecell{$\color{blue}{\lambda=0}$} \\
		\hline
		\makecell{Tip-\\Adapter\\\cite{TipAdapter}}& \cellcolor{LightRed} \makecell{$\text{L2}(\text{P}(\text{E}_\text{v}(I)))$\\$\theta_\text{im}=\varnothing$} &\cellcolor{LightGreen} \makecell{$\color{blue}{n_1=CM,n_2=C}$\\$\color{blue}{A_i^\text{im}=\text{L2}(\text{P}(\text{E}_\text{v}(I_i))), i=1,\cdots,n_1}$\\$\color{blue}{A_c^\text{txt}=\text{L2}(\text{P}(\text{E}_\text{t}(\mathcal{C}_c))), c=1,\cdots,n_2}$\\${\theta_\text{a}=\varnothing}$}&\cellcolor{LightBlue} \makecell{$\color{blue}{k_i\!=\!k_\text{gau}}$\\$\color{blue}{i=1,\cdots,n_1}$\\$\color{blue}{k_i\!=\!k_\text{lin}}$\\$\color{blue}{i\in[n_1, n_1\!+\!n_2]}$\\${\theta_\text{ker}=\varnothing}$} &  \cellcolor{LightIndigo} \makecell{$\color{blue}{ V(\theta_\text{log})\!=\!\left[\alpha Z,\mathbb{I}\right]}$\\$\theta_\text{log}=\varnothing$} &\cellcolor{almond}\makecell{N/A}\\
		\hline
		\makecell{Tip-\\Adapter-F\\\cite{TipAdapter}}& \cellcolor{LightRed} \makecell{$\text{L2}(\text{P}(\text{E}_\text{v}(I)))$\\$\theta_\text{im}=\varnothing$} & \cellcolor{LightGreen} \makecell{\color{blue}{$n_1=CM,n_2=C$}\\$\color{blue}{A_c^\text{txt}=\text{L2}(\text{P}(\text{E}_\text{t}(\mathcal{C}_c)))}$\\$\color{blue}{\theta_\text{a}=\mathcal{A}_\text{im}}$}& \cellcolor{LightBlue}\makecell{$\color{blue}{k_i\!=\!k_\text{gau}}$\\$\color{blue}{i=1,\cdots, n_1}$\\$\color{blue}{k_i\!=\!k_\text{lin}}$\\$\color{blue}{i\in[n_1,n_1\!+\!n_2]}$\\${\theta_\text{ker}=\varnothing}$} &  \cellcolor{LightIndigo} \makecell{$\color{blue}{ V(\theta_\text{log})\!=\!\left[\alpha Z,\mathbb{I}\right]}$\\$\theta_\text{log}=\varnothing$} &\cellcolor{blond}\makecell{$\color{blue}{\lambda=0}$}\\
		\hline
		\makecell{SuS-X\\\cite{SuS}}& \cellcolor{LightRed} \makecell{$\text{L2}(\text{P}(\text{E}_\text{v}(I)))$\\$\theta_\text{im}=\varnothing$} &\cellcolor{LightGreen} \makecell{$\color{blue}{n_1=CM,n_2=C}$\\$\color{blue}{A_i^\text{im}=\text{L2}(\text{P}(\text{E}_\text{v}(I_i^\text{syn}))), i=1,\cdots,n_1}$\\$\color{blue}{A_c^\text{txt}=\text{L2}(\text{P}(\text{E}_\text{t}(\mathcal{C}_c))), c=1,\cdots,n_2}$\\${\theta_\text{a}=\varnothing}$}&\cellcolor{LightBlue} \makecell{$\color{blue}{k_i\!=\!k_\text{calib-gau}}$\\$\color{blue}{i=1,\cdots, n_1}$\\$\color{blue}{k_i\!=\!k_\text{lin}}$\\$\color{blue}{i\in[n_1, n_1\!+\!n_2]}$\\${\theta_\text{ker}=\varnothing}$} &  \cellcolor{LightIndigo} \makecell{$\color{blue}{ V(\theta_\text{log})\!=\!\left[ Z,\mathbb{I}\right]}$\\$\theta_\text{log}=\varnothing$} &\cellcolor{almond}\makecell{N/A}\\
		\hline
		\makecell{CPR\cite{CPR}}& \cellcolor{LightRed} \makecell{$\text{L2}(\text{P}(\text{E}_\text{v}(I)))$\\$\theta_\text{im}=\varnothing$} & \cellcolor{LightGreen}\makecell{$\color{blue}{n_1=C, n_2=C}$\\$\color{blue}{A_i^\text{im}=\frac{1}{m}\sum\limits_{t\in \mathcal{I}_i}\text{E}_\text{v}(I_t), i=1,\cdots,n_1}$\\$\color{blue}{A_c^\text{txt}=\text{E}_\text{t}(\mathcal{C}_c)+B_c^\text{txt}+B_c^\text{im}}$\\$\color{blue}{\theta_\text{a}=\theta_1\cup\theta_2}$} &\cellcolor{LightCyan} \makecell{${k_i\!=\!k_\text{lin}}$\\${i=1,\cdots,n_1\!+\!n_2}$\\${\theta_\text{ker}=\varnothing}$}& \cellcolor{LightMagenta} \makecell{$\color{blue}{ V(\theta_\text{log})\!=\!\left[(1\!-\!\alpha){\Gamma},\alpha{\Gamma}\right]}$\\$\theta_\text{log}=\varnothing$}&\cellcolor{babypink}\makecell{$\color{blue}{\lambda>0}$}\\
		\hline
		\makecell{APE\cite{APE}}&\cellcolor{LightRed} \makecell{$\text{L2}(\text{P}(\text{E}_\text{v}(I)))$\\$\theta_\text{im}=\varnothing$}&\cellcolor{LightGreen} \makecell{$\color{blue}{n_1=CM,n_2=C}$\\$\color{blue}{A_i^\text{im}=\text{L2}(\text{P}(\text{E}_\text{v}(I_i))),i=1,\cdots,n_1}$\\$\color{blue}{A_c^\text{txt}=\text{L2}(\text{P}(\text{E}_\text{t}(\mathcal{C}_c))),c=1,\cdots,n_2}$\\${\theta_\text{a}=\varnothing}$}& \cellcolor{LightBlue}\makecell{$\color{blue}{k_i\!=\!k_\text{gau}}$\\$\color{blue}{i=1,\cdots, n_1}$\\$\color{blue}{k_i\!=\!k_\text{lin}}$\\$\color{blue}{i\in[n_1,n_1\!+\!n_2]}$\\${\theta_\text{ker}=\varnothing}$} &  \cellcolor{LightIndigo} \makecell{$\color{blue}{ V(\theta_\text{log})\!=\!\left[\alpha Z\text{diag}(R_\text{FW}),\mathbb{I}\right]}$\\$\theta_\text{log}=\varnothing$}& \cellcolor{almond}\makecell{N/A}\\
		\hline
		\makecell{APE-T\\\cite{APE}}&\cellcolor{LightRed} \makecell{$\text{L2}(\text{P}(\text{E}_\text{v}(I)))$\\$\theta_\text{im}=\varnothing$}&\cellcolor{LightGreen} \makecell{$\color{blue}{n_1=CM,n_2=C}$\\$\color{blue}{A_i^\text{im}\!=\!\text{L2}(\text{P}(\text{E}_\text{v}(I_i))\!+\!B_{\lceil i/C\rceil})}, i=1,\cdots,n_1$\\$\color{blue}{A_c^\text{txt}=\text{L2}(\text{P}(\text{E}_\text{t}(\mathcal{C}_c))+B_c)}, c=1,\!\cdots\!,\!n_2$\\$\color{blue}{\theta_\text{a}=\{B\}}$}&\cellcolor{LightBlue} \makecell{$\color{blue}{k_i\!=\!k_\text{gau}}$\\$\color{blue}{i=1,\!\cdots\!,\!n_1}$\\$\color{blue}{k_i\!=\!k_\text{lin}}$\\$\color{blue}{i=n_1,\cdots,n_1\!+\!n_2}$\\${\theta_\text{ker}=\varnothing}$} &  \cellcolor{LightMagenta} \makecell{$\color{blue}{ V(\theta_\text{log})\!=\!\left[\alpha Z\text{diag}(R_\text{FW}),\mathbb{I}\right]}$\\$\color{blue}{\theta_\text{log}=\{R_\text{FW}\}}$}& \cellcolor{blond}\makecell{$\color{blue}{\lambda=0}$}\\
		\hline
	\end{tabular}%
	\addtolength{\tabcolsep}{0.7em}
	\label{tab:compare}
\end{table*}

\begin{figure*}
	\centering
	\includegraphics[width=0.95\linewidth]{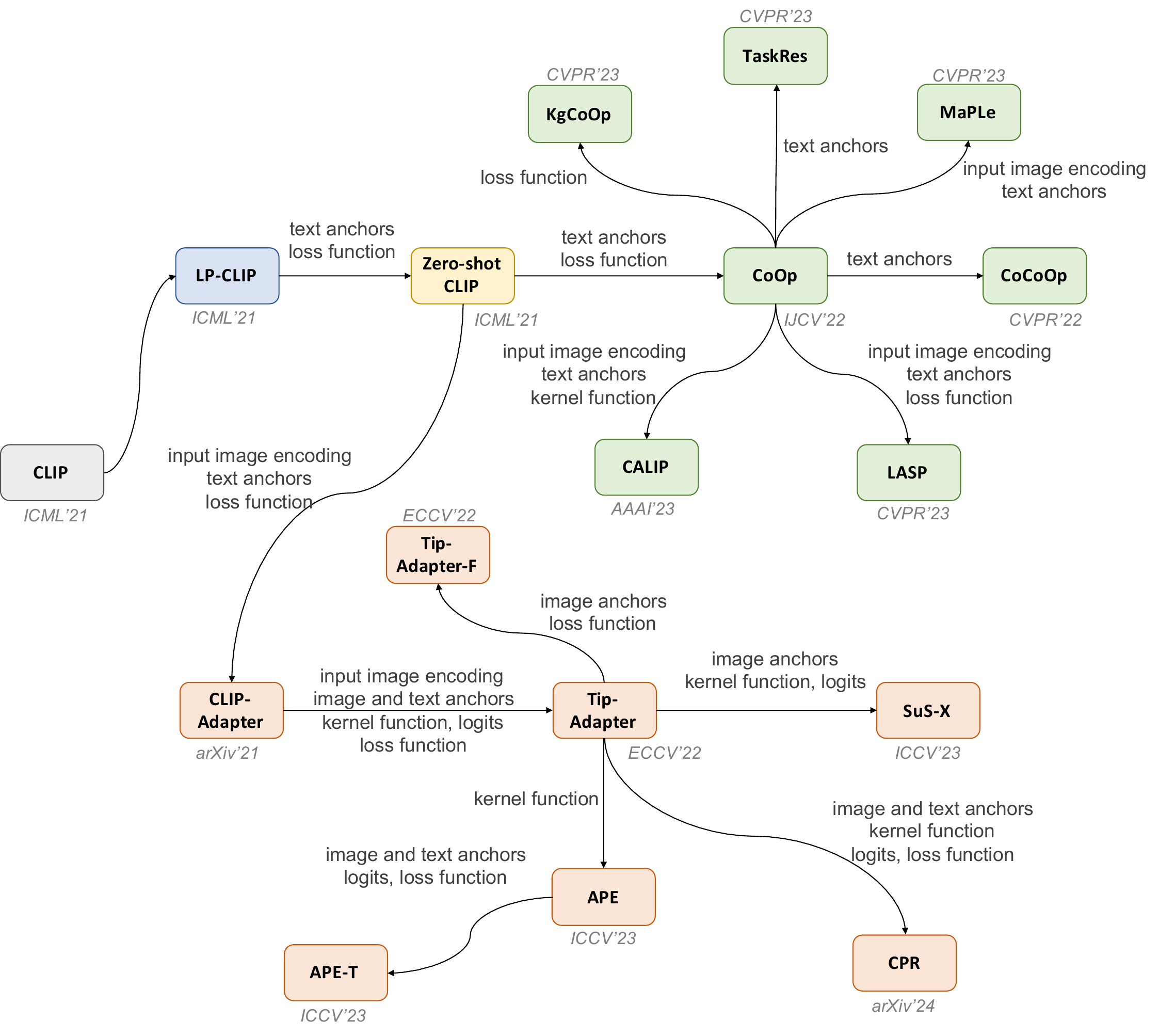}
	\caption{Evolution of different parameter-efficient adaptation methods of CLIP. The texts around arrows represent the differences between source and destination method.}
	\label{fig:evolution_graph}
\end{figure*}

\subsubsection{APE and APE-T}
APE~\cite{APE} uses CLIP's image encoder to extract image features, i.e., $X=\phi_\text{im}(I; \theta_\text{im})=\text{L2}(\text{P}(\text{E}_\text{v}(I)))$ with $\theta_\text{im}=\varnothing$. The $CM$ image anchors of all labeled training samples are computed by $A_i^\text{im}=\text{L2}(\text{P}(\text{E}_\text{v}(I_i)))$ for $i=1,\cdots,n_1$ with $n_1=CM$. The text anchors are obtained by $A_c^\text{im}=\text{L2}(\text{P}(\text{E}_\text{t}(\mathcal{C}_c)))$ for $c=1,\cdots, n_2$ with $n_2=C$. Thus, we have $\theta_\text{a}=\varnothing$.

APE improves Tip-Adapter by introducing the similarities between training image features and text features as the scores of training image features in the cache model. The scores are computed by
\begin{align}
{R}_\text{FW}&=\exp\big(\gamma \text{KL}(W^TF'| Z)\big)\in\mathbb{R}^{n_1},\\
F'&=[A_1^\text{im},\cdots, A_{n_1}^\text{im}]^T,\\
W&=[A_1^\text{txt},\cdots, A_{n_2^\text{txt}}]^T,
\end{align}
where $\gamma$ is a hyper-parameter and $\text{KL}$ denotes the KL-divergence between CLIP's prediction and one-hot labels. Integrating logits from text features and training image features, the overall logits are:
\begin{align}
\text{logits}=W^TX+\alpha  Z\text{diag}(R_\text{FW})R_\text{fF}.
\end{align}
In this equation, $\alpha$ denotes balance factor and $\text{diag}()$ is diagonalization. Besides, $R_\text{fF}$ denotes the affinities between $X$ and $F'$:
\begin{align}
R_\text{fF}=\exp\big(-\beta(1-(F')^TX)\big)\in\mathbb{R}^{n_1\times 1},
\end{align}
where $\beta$ denotes kernel parameter. The logits can be reformulated into
\begin{align}
f(X)&=\big[\alpha Z\text{diag}(R_\text{FW}),\mathbb{I}\big]\begin{bmatrix}k_\text{im}(X)\\k_\text{txt}(X)\end{bmatrix}\nonumber\\
&\triangleq V(\theta_\text{log})\cdot k(X),
\end{align}
where 
\begin{align}
k_\text{im}(X)=\big[k_1(X, A_1^\text{im}), \cdots, k_{n_1}(X, A_{n_1}^\text{im})\big]^T
\end{align}
with $k_i=k_\text{gau}$ ($i=1,\cdots,n_1$) and 
\begin{align}
k_\text{txt}(X)=\big[k_{n_1+1}(X, A_1^\text{txt}), \cdots, k_{n_1\!+\!n_2}(X, A^\text{txt}_{n_2})\big]^T
\end{align} 
with $k_c=k_\text{lin}$ ($c=n_1+1,\dots, n_1+n_2$). The kernels are not learnable, therefore we have $\theta_\text{ker}=\varnothing$.

APE-T~\cite{APE} is the training-required version of APE. APE-T freezes the cache model and only trains lightweight category residuals $B=[B_1,\cdots,B_{n_2}]\in\mathbb{R}^{d\times n_2}$ as well as the cache scores $R_\text{FW}\in\mathbb{R}^{n_1}$. The image anchors are
\begin{align}
A_i^\text{im}=\text{L2}\big(\text{P}(\text{E}_\text{v}(I_i)) + B_{\lceil i/n_2\rceil}\big), i=1,\cdots, n_1,
\end{align}
where $\lceil\cdot\rceil$ denotes the ceiling function. The text anchors are
\begin{align}
A_c^\text{txt}=\text{L2}\big(\text{P}(\text{E}_\text{t}(\mathcal{C}_c))+B_c\big), c=1,\cdots, n_2.
\end{align}
Thus, the parameter of $\phi_\text{a}$ is $\theta_\text{a}=\{B\}$. $ V(\theta_\text{log})$ is learnable now, thus, $\theta_\text{log}=\{R_\text{FW}\}$. APE-T uses the same training objective to CoOp and the learnable parameters are $\theta=\{\theta_\text{im}, \theta_\text{a}, \theta_\text{ker},\theta_\text{log}\}=\{B, R_\text{FW}\}$.

\subsection{Comparison of Existing Methods}
Based on the above analyses, different PEFT methods are intuitively compared in Table~\ref{tab:compare} from five dimensions. The similarities and differences of different methods are discussed in the following.

In terms of input image encoding, LP-CLIP, zero-shot CLIP, CoOp, CoCoOp, KgCoOp, TaskRes, CALIP, Tip-Adapter, Tip-Adapter-F, SuS-X, CPR, APE and APE-T use frozen CLIP's image encoder to extract input image's features. While MaPLe, LASP and CLIP-Adapter introduce learnable parameters to adapt image features to new training samples.

In terms of anchors computation, LP-CLIP, zero-shot CLIP, CoOp, CoCoOp, KgCoOp, TaskRes, MaPLe, CALIP, LASP, CLIP-Adapter only use text anchors. By contrast, Tip-Adapter, Tip-Adapter-F, SuS-X, CPR, APE and APE-T additionally define image anchors that serve keys in cache model. Among these methods, CPR only uses $C$ image anchors, while the rest use $CM$ image anchors.

In terms of kernels computation, LP-CLIP, zero-shot CLIP, CoOp, CoCoOp, KgCoOp, TaskRes, MaPLe, CALIP, LASP, CLIP-Adapter and CPR use linear kernel, while the rest use non-linear kernels. The Gaussian kernel is the most commonly used non-linear kernel. We can also find that, except the parameters in image features and anchors, no other learnable parameters are introduced while computing kernels in all of the surveyed methods.

In terms of logits computation, LP-CLIP, zero-shot CLIP, CoOp, CoCoOp, KgCoOp, TaskRes, MaPLe, CALIP, LASP and CLIP-Adapter use the identity matrix as the transformation matrix. Among the rest methods, all methods except CPR use the label matrix $Z$ of training samples to define more complex transformation matrix. In all of the listed method, except APE-T and CPR, non-learnable transformation matrix is adopted. Last but not least, as will be explained later, all of the compared methods regard the dimensions belonging to different classes in $k(X)$ to be independent.

In terms of loss function, only KgCoOp and LASP in prompt-based methods adopt regularization term. While in adapter-based methods, only CPR introduces regularization term.

The evolution graph of different methods are visualized in Figure~\ref{fig:evolution_graph}, where the dimensions of difference between source and destination methods are annotated. Obviously, two groups of methods can be identified: prompt-based and adapter-based methods. Different prompt-based methods are mainly derived from CoOp, while adapter-based methods are mainly derived from CLIP-Adapter. Based on the above comparisons, in the following sections, we will first present some possible variants within our framework and then demonstrate an implementation that extends the transformation matrix in logits computation. After that, we validate the effectiveness by extensive experiments on 11 datasets.

\section{Possible Variants}
\label{sec:imprv_tech}
Based on the above summarization, we propose several possible variants (summarized in Table~\ref{tab:imprv}) to further improve existing works. Most of them can be derived naturally and included in our framework. However, these variants have rarely been explored in previous works.

\subsection{Inter-Class Correlation}
All compared methods treat the dimensions belonging to different classes of $k(X)$ to be independent while computing logits. In other words, the representers $\{k_i(X_i, \cdot)\}_{i=1}^{n}$ in RKHS are regarded to be independent. For example, in Tip-Adapter, the final logits are computed by $f(X)=[\alpha  Z,\mathbb{I}]k(X)$, which can be converted into
\begin{align}
f(X)&=\alpha\Big[\sum\limits_{i=1}^K k(X, A_i),\cdots, \sum\limits_{i=(C-1)K+1}^{CM}k(X, A_i)\Big]^T\nonumber\\
&\qquad + \Big[k(X,A_{CM+1}), \cdots, k(X, A_{CM+C})\Big]^T.
\end{align}
From this equation, we can see that the logit for each class only pools the information in $k(X)$ for this class. Although the features $\{k_i(X, A_i)\}_{i=1}^n$ already lie in high-level space, they should be correlated each other still. Therefore, one possible direction to further extend existing methods is exploiting inter-class information while computing logits, that is modeling inter-class correlation between values of $k(X)$. This could be achieved by defining more complex transformation matrix $ V(\theta_\text{log})$.

\begin{table}[!thb]
	\centering
	\caption{Summary of possible variants within our framework.}
	\renewcommand{\arraystretch}{1.3}
	\addtolength{\tabcolsep}{0.1em}
	\begin{tabular}{ll}
		\toprule
		\textbf{Dimension} & \textbf{Possible Variants}\\
		\midrule
		%input image encoding &\\
		anchors computation & Interaction between Anchors\\
		\multirow{2}{*}{kernels computation} &More Complex Kernels\\
		&Multi-Kernel Learning\\
		\multirow{2}{*}{logits computation} & Inter-class Correlation\\
		& Learnable Transformation Matrix\\
		loss function computation & Regularization \\
		\bottomrule
	\end{tabular}
	\addtolength{\tabcolsep}{-0.1em}
	\label{tab:imprv}
\end{table}

\subsection{Learnable Transformation Matrix}
As previously stated, most of the existing methods define a fixed transformation matrix $ V(\theta_\text{log})$. However, we believe that this transformation matrix could attain an enhanced level of learnability while simultaneously evading overfitting. This could be achieved by constraining it to have a parametric form, making its values be dependent only on a few learnable parameters. Notably, APE~\cite{APE} has conducted preliminary explorations in this aspect, employing a clear and concise method for column-wise weighting.

\subsection{More Complex Kernels}
As can be seen from Table~\ref{tab:compare}, most of the existing methods define kernels over pooled image and text features. The pooling operations cause the loss of two-dimensional structural information in images, and the loss of temporal information in text, possibly limiting the ability for fine-grained recognition. To overcome this shortcoming, one promising direction is defining more complex kernels over the unpooled image and text features. In the kernels, fine-grained token-to-token similarity can be reflected, strengthening the key parts of local image and language information.

\subsection{Multi-Kernel Learning}
By investigating the existing works, we find that the kernel function $k_i$ in most of the existing works is simply a single pre-defined kernel, for example, linear kernel or Gaussian kernel. Actually, the composite kernel that combines multiple kernels in a linear/non-linear or data-dependent manner~\cite{MKL} can also be integrated into our framework. By combining information from multiple sources, the model is expected to have stronger representation ability and lesser sensitivity to kernel parameters.

\subsection{Interaction between Anchors}
In most of the existing methods, the image anchors and text anchors are usually respectively learned without more in-depth interaction. With the information from few-shot training data, it is possible to further dig text-related image features and image-related text features. This information is expected to enhance models' ability on specific downstream tasks. It is worth nothing that it is important to design parameter-efficient structure to avoid introducing too many parameters and too much computation burden.

\subsection{Regularization}
The typical training objective in existing works is image-text contrastive loss. However, using the loss only is demonstrated to generate models overfitting on training set~\cite{CoOp}. Introducing more dedicated regularization term, for example the constraints by distilling knowledge from models with stronger generalization ability~\cite{KgCoOp,LASP}, is expected to alleviate this problem.

\begin{table*}[t]
	\centering
	\caption{Results of different prompt-based methods (16 shots). $\Delta$ denotes the gap between a method ``A+*" and A. The results of MaPLe with ResNet-50 backbone are not given as its source code does not support this backbone.}
	\label{tab:prompt_results}
	\renewcommand{\arraystretch}{1.28}
	\addtolength{\tabcolsep}{0.2em}
	\begin{tabular}{l cccccccccccc}
		\toprule
		& \rotbox{\textbf{ImageNet}} & \rotbox{\textbf{Caltech101}} & \rotbox{\textbf{OxfordPets}} & \rotbox{\textbf{StanfordCars}} & \rotbox{\textbf{Flowers102}} & \rotbox{\textbf{Food101}} & \rotbox{\textbf{FGVCAircraft}} & \rotbox{\textbf{SUN397}} & \rotbox{\textbf{DTD}} & \rotbox{\textbf{EuroSAT}} & \rotbox{\textbf{UCF101}} & \rotbox{\textbf{Average}}\\
		\toprule
		\multicolumn{5}{l}{\textbf{Vision backbone: ViT-B/16}}\\
		ZS-CLIP & 66.71 & 90.19 & 88.90 & 65.30 & 70.19 & 85.96 & 24.85 & 62.61 & 44.21 & 47.73 & 66.75 & 64.85 \\
		CoOp	&	72.00 &	95.63 &	93.23	& 80.97 &	96.27 &	85.97 &	40.77&	75.27&	69.30&	81.27&	82.07&	79.34\\
		CoOp+cache	&	65.90&	94.63&	87.43&	76.80&	95.50&	81.53&	43.17&	71.27&	67.73&	76.00&	78.70 & 76.24\\
		\rowcolor{tabhighlight}
		$\Delta$ & -6.10	& -1.00	& -5.80	& -4.17	& -0.77	& -4.44	& +2.40&	-4.00&	-1.57&	-5.27&	-3.37&	-3.10 \\
		CoOp+KRR&	73.13&	95.87&	91.67&	84.07&	97.97&	86.30&	49.63&	76.33&	72.73&	83.73&	84.90& 81.48\\
		\rowcolor{tabhighlight}
		$\Delta$&+1.13&	+0.24&	-1.56&	+3.10&	+1.70&	+0.33&	+8.86&	+1.06&	+3.43&	+2.46&	+2.83&	+2.14\\
		\midrule
		\multicolumn{5}{l}{\textbf{Vision backbone: ViT-B/16}}\\
		KgCoOp&		71.37&	95.00&	93.03&	75.87&	93.43&	87.13&	35.73&	74.47&	68.57&	74.33&	80.63&	77.23\\
		KgCoOp+cache &67.73&			94.40 &	88.17&	75.27&	95.40&	84.17&	43.63&	73.27&	67.30&	75.40&	78.70 & 76.68 \\
		\rowcolor{tabhighlight}
		$\Delta$ &-3.64& -0.60 &	-4.86	& -0.60 &	+1.97 &	-2.96 &	+7.90 &	-1.20 &	-1.27 &	+1.07 &	-1.93 & -0.56\\
		KgCoOp+KRR	&73.43& 96.27	&92.80&	83.30&	97.40&	87.60&	48.97&	76.90&	73.60&	84.10&	85.37 &81.79 \\
		\rowcolor{tabhighlight}
		$\Delta$ & +2.06&+1.27&	-0.23&	+7.43&	+3.97&	+0.47&	+13.24&	+2.43&	+5.03&	+9.77&	+4.74 & +4.56\\
		\midrule
		\multicolumn{5}{l}{\textbf{Vision backbone: ViT-B/16}}\\
		MaPLe&71.93&			95.33&	93.60 &	74.27&	93.47&	87.47&	37.27&	74.43&	67.23&	85.67&	81.27 & 78.36\\
		MaPLe+cache &		70.60&	94.90&	90.20&	78.73&	96.67&	84.50&	42.70&	74.27&	69.50&	73.30&	79.90& 77.75\\
		\rowcolor{tabhighlight}
		$\Delta$ &-1.33& -0.43&	-3.40&	+4.46&	+3.20&	-2.97&	+5.43&	-0.16&	+2.27&	-12.37&	-1.37 & -0.61\\
		MaPLe+KRR	&73.40&	96.10 &	92.10 &	84.13&	97.93&	87.17&	49.10&	76.67&	72.93&	83.83&	85.77 & 81.74\\
		\rowcolor{tabhighlight}
		$\Delta$ & +1.47& +0.77	&-1.50	&+9.86&	+4.46&	-0.30&	+11.83&	+2.24&	+5.70&	-1.84&	+4.50 & +3.38\\
		\midrule
		\multicolumn{5}{l}{\textbf{Vision backbone: ResNet-50}}\\
		ZS-CLIP & 60.32	&85.72&	85.77&	55.81&	66.06&	77.39&	17.01&	58.84&	42.97&	36.19&	61.80&	58.90\\
		CoOp&63.03&	91.90&	88.33&	71.27&	93.60	&76.97&	29.33&	69.97&	64.24	&78.60&	76.97&73.11\\
		CoOp+cache &56.25& 90.37	&79.53&	65.47&	91.83&	70.70&	31.53&	63.93	&62.13	&75.70	&70.60&68.91\\
		\rowcolor{tabhighlight}
		$\Delta$ &-6.78& -1.53&	-8.80&	-5.80&	-1.77&	-6.27&	+2.20&	-6.04&	-2.11&	-2.90&	-6.37&-4.20\\
		CoOp+KRR &63.80& 92.63&	85.93	&75.87&	96.23&	77.30	&39.23	&71.20	&67.73&	81.70&	78.87&75.50\\
		\rowcolor{tabhighlight}
		$\Delta$ &+0.77& +0.73&	-2.40&+4.60	&+2.63&	+0.33&	+9.90	&+1.23&	+3.49	&+3.10	&+1.90&+2.39\\
		\midrule
		\multicolumn{5}{l}{\textbf{Vision backbone: ResNet-50}}\\
		KgCoOp &63.13& 90.63	&88.57&	67.70	&89.17&	79.20	&25.63&	69.43&	64.57&	70.07&	74.07&71.11\\
		KgCoOp+cache &57.15& 90.10	&79.83	&65.13	&90.60	&72.87	&32.40	&66.87	&63.10	&73.67	&71.10&69.35\\
		\rowcolor{tabhighlight}
		$\Delta$&-5.98&-0.53	&-8.74	&-2.57&	+1.43&	-6.33&	+6.77&	-2.56&	-1.47&	+3.60&	-2.97&-1.76\\
		KgCoOp+KRR &64.97& 93.30&	88.37&	75.03	&95.53	&79.37&	39.30	&72.07	&68.67	&81.10&	78.87 &76.05 \\
		\rowcolor{tabhighlight}
		$\Delta$ &+1.84& +2.67&	-0.20&	+7.33&	+6.36&	+0.17&	+13.67&	+2.64&	+4.10&	+11.03&	+4.80& +4.95\\
		\bottomrule
	\end{tabular}
	\addtolength{\tabcolsep}{-0.2em}
\end{table*}

\begin{table*}[t]
	\centering
	\caption{Results of different adapter-based methods (1$\sim$16 shots). $\Delta$ means the difference between Tip-Adapter+KRR and Tip-Adapter or between Tip-Adapter-F+KRR and Tip-Adapter-F.}
	\label{tab:adapter_results}
	%\footnotesize
	\renewcommand{\arraystretch}{1.28}
	\addtolength{\tabcolsep}{-0.1em}
	\begin{tabular}{l ccccccccccccc}
		\toprule
		&\rotbox{\textbf{Setting}} & \rotbox{\textbf{ImageNet}} & \rotbox{\textbf{Caltech101}} & \rotbox{\textbf{OxfordPets}} & \rotbox{\textbf{StanfordCars}} & \rotbox{\textbf{Flowers102}} & \rotbox{\textbf{Food101}} & \rotbox{\textbf{FGVCAircraft}} & \rotbox{\textbf{SUN397}} & \rotbox{\textbf{DTD}} & \rotbox{\textbf{EuroSAT}} & \rotbox{\textbf{UCF101}} & \rotbox{\textbf{Average}}\\
		\toprule
		zero-shot CLIP& 0-shot&		60.32&	85.72&	85.77&	55.81&	66.06&	77.39&	17.01&	58.84&	42.97&	36.19&	61.80&	58.90\\
		\midrule
		CLIP-Adapter&&61.20&	88.60&	85.99&	55.13&	73.49&	76.82&	17.49&	61.30&	45.80&	61.40&	62.20&	62.67\\
		Tip-Adapter&&		60.47&	87.71&	86.02&	57.27&	73.81&	77.44&	18.99&	61.32&	46.22&	49.60&	63.73&	62.05\\
		Tip-Adapter+KRR&& 60.87	&88.56	&86.07	&58.03	&76.61	&77.50&	20.40&	61.57&	47.81&	51.77&	63.71&	62.99\\
		\rowcolor{tabhighlight}
		$\Delta$ && +0.40&	+0.85&	+0.05&	+0.76&	+2.80&	+0.06&	+1.41&	+0.25&	+1.59&	+2.17&	-0.02&	+0.94 \\
		Tip-Adapter-F	&&	60.96&	89.49&	86.43&	58.90&	79.29	&77.65	&20.64&	62.50&	49.53&	56.67&	65.32&	64.31\\
		Tip-Adapter-F+KRR&&	61.15&	89.25&	87.08&	59.59&	77.51&	77.62&	21.03&	62.28&	49.11&	54.74&	65.58&	64.09\\
		\rowcolor{tabhighlight}
		$\Delta$ &\multirow{-7}{*}{1-shot}& +0.19 &	-0.24	& +0.65	& +0.69&	-1.78&	-0.03&	+0.39&	-0.22&	-0.42&	-1.93&	+0.26&	-0.22 \\
		\midrule
		CLIP-Adapter&&		61.52&	89.37&	86.73&	58.74	&81.61	&77.22	&20.10&	63.29&	51.48&	63.90&	67.12&	65.55\\
		Tip-Adapter	&&	60.96&	88.44	&87.03&	57.93&	79.13	&77.52&21.21&	62.70&	49.47&	61.68&	64.74&	64.62\\
		Tip-Adapter+KRR&&61.22&	89.29&	87.05&	60.74&	83.23	&77.68	&23.58&	63.87&	51.77&	60.49	&65.82	&65.89\\
		\rowcolor{tabhighlight}
		$\Delta$ && +0.26&	+0.85&	+0.02&	+2.81&	+4.10&	+0.16&	+2.37&	+1.17&	+2.30&	-1.19&	+1.08&	+1.27 \\
		Tip-Adapter-F&&		61.69&	89.74	&87.03	&61.50&	82.30&	77.81&	23.19&	63.64	&53.72&	66.15&	66.43&	66.65\\
		Tip-Adapter-F+KRR&&	61.79&	89.49&	87.49&	62.33	&83.64	&77.74&	23.58&	64.79&	52.25&	65.07&	68.60&	66.98\\
		\rowcolor{tabhighlight}
		$\Delta$ &\multirow{-7}{*}{2-shot}& +0.10&	-0.25&	+0.46&	+0.83&	+1.34&	-0.07&	+0.39&	+1.15&	-1.47&	-1.08&	+2.17&	+0.32\\
		\midrule
		CLIP-Adapter&	&	61.84&	89.98&	87.46&	62.45&	87.17&	77.92&	22.59&	65.96&	56.86&	73.38&	69.05&	68.61\\
		Tip-Adapter	&&	60.98&	89.39&	86.45&	61.45&	83.80&	77.54	&22.41	&64.15&	53.96&	65.32&	66.46&	66.54\\
		Tip-Adapter+KRR&& 61.58	&90.71&	87.33&	65.35&	90.34&	78.09&	26.97&	66.02&	59.34&	74.00&	67.20&	69.72\\
		\rowcolor{tabhighlight}
		$\Delta$ && +0.60	& +1.32&	+0.88&	+3.90&	+6.54&	+0.55&	+4.56&	+1.87&	+5.38&	+8.68&	+0.74&	+3.18 \\
		Tip-Adapter-F	&&	62.52&	90.56&	87.54&	64.57&	88.83&	78.24	&25.80&	66.21&	57.39	&74.12&	70.55&	69.67\\
		Tip-Adapter-F+KRR &&62.33&	91.20&	88.06&	66.46&	89.44&	78.36&	25.26&	67.46&	60.22&	78.31&	71.27&	70.76\\
		\rowcolor{tabhighlight}
		$\Delta$ &\multirow{-7}{*}{4-shot}& -0.19	&+0.64&	+0.52&	+1.89&	+0.61&	+0.12&	-0.54&	+1.25&	+2.83&	+4.19	&+0.72&	+1.09\\
		\midrule
		CLIP-Adapter&	&	62.68&	91.40&	87.65&	67.89&	91.72&	78.04&	26.25&	67.50&	61.00&	77.93	&73.30&	71.40\\
		Tip-Adapter	&&	61.45&	89.83&	87.03&	62.93&	87.98&	77.76	&25.59&	65.62&	58.63&	67.95&	68.68&	68.50\\
		Tip-Adapter+KRR&&	62.16&	91.40&	88.31&	69.42	&94.48	&78.54&	33.66&	68.48	&63.83	&81.06	&68.83	&72.74\\
		\rowcolor{tabhighlight}
		$\Delta$&& +0.71	& +1.57	&+1.28&	+6.49&	+6.50&	+0.78&	+8.07&	+2.86&	+5.20&	+13.11&	+0.15&	+4.25 \\
		Tip-Adapter-F&&		64.00&	91.44&	88.09&	69.25&	91.51&	78.64&	30.21&	68.87&	62.71	&77.93&	74.25	&72.45\\
		Tip-Adapter-F+KRR&&	63.55&	92.21&	88.83&	69.87	&94.56&	78.88&	31.41&	69.23&	62.65&	82.07&	74.65&	73.45\\
		\rowcolor{tabhighlight}
		$\Delta$ &\multirow{-7}{*}{8-shot}& -0.45&	+0.77&	+0.74&	+0.62&	+3.05&	+0.24&	+1.20	&+0.36&	-0.06&	+4.14&	+0.40&	+1.00\\
		\midrule
		CLIP-Adapter& &		63.59&	92.49&	87.84&	74.01&	93.90&	78.25&	32.10&	69.55&	65.96&	84.43&	76.76&	74.44\\
		Tip-Adapter	& &	62.01&	90.18&	88.14&	66.77&	89.89&	77.83&	29.76&	66.85&	60.93&	70.54&	70.58&	70.32\\
		Tip-Adapter+KRR& &	62.67&	92.49&	89.42&	75.74&	95.90&	79.33&	38.88&	70.69&	67.73&	81.00&	70.39&	74.93\\
		\rowcolor{tabhighlight}
		$\Delta$&&+0.66	&+2.31	&+1.28&	+8.97&	+6.01&	+1.50	&+9.12&	+3.84&	+6.80&	+10.46&	-0.19&	+4.61\\
		Tip-Adapter-F&&		65.51&	92.86&	89.70&	75.48&	94.80&	79.43&	35.55&	71.47&	66.55&	84.54&	78.03&	75.81\\
		Tip-Adapter-F+KRR&&	64.86&	93.63&	89.45&	75.96&	96.39&	79.26&	37.14&	71.50&	68.09&	82.68&	79.20&	76.20\\
		\rowcolor{tabhighlight}
		$\Delta$ &\multirow{-7}{*}{16-shot}& -0.65	& +0.77&	-0.25&	+0.48&	+1.59	&-0.17&	+1.59&	+0.03&	+1.54&	-1.86&	+1.17&	+0.39\\
		\bottomrule
	\end{tabular}
	\addtolength{\tabcolsep}{0.1em}
\end{table*}

\section{A Demonstrative Implementation}
\label{sec:extension}
As mentioned above, all surveyed methods do not consider the inter-class correlation between different dimensions of $k(X)$. This section will demonstrate a method to extend the transformation matrix while modeling the inter-class correlation. 

Our method is quite simple, which is inspired by the relation between cache model and KRR. In cache model, the query is $X$, the keys are $X_1, \cdots, X_n$, and the values are collected by the label matrix $Z$. The output of the cache model is
\begin{align}
X_\text{cache}=Z k(X),
\end{align}
where $k(X)=[k(X, X_1), \cdots, k(X, X_n)]^T$. In cache model, the Gaussian kernel is used for computing $k(X)$. In the multivariant case, the prediction of KRR is
\begin{align}
X_\text{KRR}=Z(K+\lambda \mathbb{I})^{-1}k(X),
\end{align}
where $K$ is kernel matrix. Interestingly, the cache model can be seen as a simplified version of KRR, where $(K+\lambda \mathbb{I})^{-1}$ is replaced by $\mathbb{I}$. Compared to cache model, the correlation between different dimensions of $k(X)$ can be reflected by $(K+\lambda \mathbb{I})^{-1}$. Due to such a connection, the cache model in existing methods can be replaced by the solution of KRR. For example, the new transformation matrix in logits computation of Tip-Adapter-F can be represented as
\begin{align}
V(\theta_\text{log})=\begin{bmatrix} \alpha Z(K+\lambda \mathbb{I})^{-1}, \mathbb{I}\end{bmatrix}.
\end{align}
As for prompt-based methods, we can also add a cache model and replace it by the solution of KRR. For example, the updated transformation matrix in CoOp can be represented as
\begin{align}
V(\theta_\text{log})=\alpha Z(K+\lambda \mathbb{I})^{-1}+\mathbb{I}, 
\end{align}
where $\alpha$ is a trade-off parameter. We fuse $Z(K+\lambda \mathbb{I})^{-1}$ and $\mathbb{I}$ instead of only using the former so as to exploit their complementary.

\section{Experiments}
\label{sec:exp}
This section validates the effectiveness of using the solution of KRR to define the transformation matrix by experiments.

\textbf{Datasets and Metrics}. The experiments are conducted on 11 public datasets: ImageNet~\cite{ImageNet}, Caltech101~\cite{Caltech101}, Food101~\cite{Food101}, OxfordPets~\cite{OxfordPets}, StanfordCars~\cite{StanfordCars}, Flowers102~\cite{Flowers102}, FGVCAircraft~\cite{FGVCAircraft}, UCF101~\cite{UCF101}, EuroSAT~\cite{EuroSAT}, SUN397~\cite{SUN397}, DTD~\cite{DTD}. The same metrics to existing works are adopted for model evaluation.

\textbf{Compared Methods}. We select both prompt-based and adapter-based methods as baselines, including CoOp, KgCoOp, MaPLe, Tip-Adapter and Tip-Adapter-F. For CoOp, KgCoOp and MaPLe, the methods with cache model are implemented, which are denoted as CoOp+cache, KgCoOp+cache and MaPLe+cache, respectively. The cache models can be replaced by KRR's solution, which are denoted as CoOp+KRR, KgCoOp+KRR and MaPLe+KRR, respectively. As for Tip-Adapter and Tip-Adapter-F, when KRR's solution is used, the corresponding methods are denoted as Tip-Adapter+KRR and Tip-Adapter-F+KRR, respectively. 

\textbf{Implementation Details}. Following existing works, ViT-B/16 or ResNet-50 is used as the vision backbone for prompt-based methods, while ResNet-50 is used for adapter-based methods. Gaussian kernel is used for both cache model and KRR. In CoOp+cache, KgCoOp+cache, MaPLe+cache, CoOp+KRR, KgCoOp+KRR and MaPLe+KRR, the kernel parameter $\beta$ is set to be 5, $\alpha$ is set to be 0.01. In CoOp+KRR, KgCoOp+KRR and MaPLe+KRR, $\lambda$ is set to be 0.1. As for adapter-based methods, following previous works, the hyper-parameters are searched on validation set.

\textbf{Results of Prompt-based Methods}. The few-shot recognition results of prompt-based methods are compared in Table~\ref{tab:prompt_results}. By comparing the methods with and without using cache model, the cache model does not have positive effect on the average. By observing the $\Delta$ values on each dataset, we can find that cache model only brings positive effect on limited datasets. Specifically, when ViT-B/16 is adopted, CoOp+cache surpasses CoOp only on FGVCAircraft; KgCoOp+cache surpasses KgCoOp on Flowers102, FGVCAircraft and EuroSat; MaPLe+cache surpasses MaPLe on StanfordCars, Flowers102, FGVCAircraft and DTD. We speculate that cache model is more helpful for fine-grained tasks. While replacing cache model with the solution of KRR, improved results on the average can be achieved for CoOp, KgCoOp and MaPLe. The average accuracy boost on CoOp, KgCoOp and MaPLe is 2.14\%, 4.56\% and 3.38\%, respectively. On the datasets that cache model remarkably degrades the accuracy, the KRR's solution can rectify this situation to a great degree. The similar observations can be obtained when ResNet-50 is used. As the only difference between ``A+KRR" and ``A+cache" (A=CoOp, KgCoOp, MaPLe) is the $(K+\lambda\mathbb{I})^{-1}$ term, we can conclude that modeling the inter-class correlation between representers by this term is beneficial.

\textbf{Results of Adapter-based Methods}. The few-shot recognition results of adapter-based methods are compared in Table~\ref{tab:adapter_results}. By comparing the results of Tip-Adapter+KRR and Tip-Adapter using different shots, Tip-Adapter+KRR can achieve better results than Tip-Adapter. Overall, Tip-Adapter+KRR outperforms Tip-Adapter by 0.94\%, 1.27\%, 3.18\%, 4.25\%, 4.61\% with 1 shot, 2 shots, 4 shots, 8 shots and 16 shots, respectively. As for training-required mode, average results across all datasets of Tip-Adapter-F+KRR surpass those by Tip-Adapter-F in most cases, although not that significant compared to training-free mode. The improved results imply that modeling the inter-class correlation in adapter-based methods is also helpful.

\section{Conclusions and Discussions}
\label{sec:conclusion}
The parameter-efficient fine-tuning (PEFT) of contrastive vision-language models (CVLMs) offers a promising solution for zero- and few-shot image recognition. Various methods, both prompt-based and adapter-based, have flourished in this domain, presenting a vibrant and diverse landscape. However, these existing methods often differ significantly in terms of their underlying ideas, principles, and workflows. To support in-depth comparison of the various approaches, this paper introduces a unified computational framework grounded in the Representer Theorem, which comprises of five integral components: input image encoding, anchors computation, kernels computation, logits computation and loss function computation. By conducting comparative analyses from multiple dimensions within this framework, we propose several potential variants to improve existing methods. Beyond this, we also present a demonstrative implementation for modeling inter-class correlation between representers by exploiting the tool of kernel ridge regression. The effectiveness is validated by experiments on 11 datasets. 

Although the proposed framework can integrate many existing works and some useful variants can be found within the framework, it still has some limitations. Besides, although we have already presented some variants within our framework, we further provide more insights on the future prospects from a higher-level perspective, expanding the value of the proposed framework.

\subsection{Limitations}
First, given the vast array of methods in the literature, which often originate from diverse perspectives and make various assumptions, it is inevitable that not all of them can be encompassed within this framework. For instance, some test-time tuning methods~\cite{TTPT}, which require defining a loss function on test samples, have not been explicitly included. However, with suitable modifications, such methods could potentially be integrated. Besides, certain vision-only tuning techniques~\cite{VPT}, which are not specifically tailored for vision-language models, have also been excluded from this survey.

Second, the proposed framework focuses on adapting pre-trained contrastive vision-language models, other vision-language models such as generative ones~\cite{Flamingo,VLBEiT} are not considered. Proper adjustments on the training objectives may be needed for such models, for example, introducing the objective of masked language modeling or next-token prediction.

Third, the proposed framework is not dedicated for other low-shot vision tasks, such as low-shot image segmentation. Due to the distinct difference between dense task and classification task, special designs should be considered to achieve desirable adaptation performance~\cite{ChenDWHLDQ23}. Meantime, fine-tuning the segmentation-oriented foundation models like SAM~\cite{SAM} instead of CVLMs could be more beneficial~\cite{peng2024sam,zhong2024convolution}.

\subsection{Future Prospects}
\textbf{Pre-training with Representers}. The proposed framework is mainly for fine-tuning CVLMs for low-shot recognition task. It is worth noting that, to reduce the gap between pre-training and fine-tuning, the two procedures often adopt similar network structure and learning objective. Considering this consistency, the key idea of the proposed framework, i.e. introducing representers in RKHS, should also be suitable for pre-training CVLMs. By introducing the representers based on image anchors, it is able to learn image-image relation simultaneously, potentially enhancing the modeling capability of CVLMs and further reducing the gap between pre-training and fine-tuning. Furthermore, while defining representers, the idea of multiple kernels and dynamic kernels can also be introduced. In summary, pre-training CVLMs with representers could be a new research direction, which deserves further exploration.

\textbf{Modular and Searchable PEFT}. The proposed framework has a modular design, which includes multiple components. Each component has a high level of customizability. As such, based on the proposed framework, it is possible to implement the corresponding software framework supporting fair algorithm evaluation. Besides, in future, we could search the architectures and parameters from all components jointly by borrowing ideas from neural architecture search (NAS)~\cite{NAS}, constantly pushing the upper bound of low-shot image recognition. Different from the existing NAS methods, the data efficiency of searching algorithm should be always kept in mind in consideration of limited data.

\textbf{From Cache Model to Representer Theorem}. From this survey, existing adapter-based methods mainly consider different designs of cache model. As demonstrated in this work, cache model could be seen as a special case of the solution given by the Representer Theorem. By inspecting cache model from a higher perspective, extensive new directions can be identified. For example, by exploiting the Representer Theorem, it is possible to integrate information from multiple sources, such as text feature space generated by pre-trained LLM, image feature space generated by other kinds of pre-trained models beyond vision-language models. Besides, we can also explore some dynamic forms of representers, further expanding the capability boundary of PEFT.

\textbf{Representer Theorem and Attention Mechanism}. The Representer Theorem tells us the solution form of regularized empirical risk minimization in RKHS, which has tight relation to attention mechanism. The representers in the solution given by Representer Theorem define the similarity between queries and keys, while the transformation coefficients in prediction function can be seen as values in attention mechanism. Due to this reason, we expect the researchers can notice the potential value of Representer Theorem for advancing the architecture designing in the NLP and CV domain.

%\section*{Acknowledgments}
%This should be a simple paragraph before the References to thank those individuals and institutions who have supported your work on this article.

\bibliographystyle{IEEEtranS}
\bibliography{main}

%\vfill

\end{document}